%% file: ssvr_iclr2027.tex
\documentclass{article}
\usepackage{iclr2027_conference,times}
\usepackage{amsmath,amssymb}
\usepackage{graphicx}
\usepackage{bbding}
\usepackage{booktabs}
\usepackage{algorithm}
\usepackage{algpseudocode}
\usepackage[table]{xcolor}
\usepackage[hyphens]{url}
\usepackage{hyperref}
\graphicspath{{figures/}}

\title{Resolving State-Representation Mismatch: State-Space Visual Reasoning for Open-Loop VLA Planning}
\iclrfinalcopy 
\author{%
{\small\bfseries Junhao Xiao\textsuperscript{1}, Haoxiang Zhao\textsuperscript{2}, Menghao Fang\textsuperscript{3}, Jinkui Zhang\textsuperscript{4}, Jinghan Yu\textsuperscript{1},}\\[4pt]
{\small\bfseries Xinyu Huang\textsuperscript{1}, Zhiyu Wu\textsuperscript{1}, Kaiming Xu\textsuperscript{1}, Yi Chen\textsuperscript{4}, Youjun Bao\textsuperscript{5}, Zhiyuan Ma\textsuperscript{2,*}}\\[6pt]
{\small \textsuperscript{1}FDU\quad \textsuperscript{2}HUST\quad \textsuperscript{3}TJU\quad \textsuperscript{4}CCNU\quad \textsuperscript{5}Kuaishou}\\[4pt]
{\small \href{mailto:jhxiao26@m.fudan.edu.cn}{\texttt{jhxiao26@m.fudan.edu.cn}}\qquad \href{mailto:mzyth@hust.edu.cn}{\texttt{mzyth@hust.edu.cn}}}%
}
\ificlrfinal
\hypersetup{pdfauthor={Junhao Xiao, Haoxiang Zhao, Menghao Fang, Jinkui Zhang, Jinghan Yu, Xinyu Huang, Zhiyu Wu, Kaiming Xu, Yi Chen, Youjun Bao, Zhiyuan Ma}}
\fi

\hypersetup{
  pdftitle={Resolving State-Representation Mismatch: State-Space Visual Reasoning for Open-Loop VLA Planning},
  pdfsubject={Preprint},
  hidelinks
}

\begin{document}

\maketitle
\ificlrfinal
\begingroup
\renewcommand{\thefootnote}{\fnsymbol{footnote}}
\footnotetext[1]{Corresponding author: Zhiyuan Ma.}
\endgroup
\fi
\suppressfloats[t]

\begin{abstract}
Despite rapid progress in vision-language-action (VLA) models, existing reasoning paradigms still face a fundamental \emph{state-representation mismatch} in open-loop planning. Given only an initial observation, models must internally simulate action-conditioned state transitions, whereas text-, pixel-, and latent-space reasoning can suffer from lossy spatial compression, error-accumulating visual generation, and bypass of intermediate latent tokens, respectively, undermining reliable long-horizon planning. We propose \textbf{State-Space Visual Reasoning} (SSVR), which decouples static visual context, language constraints, and a recurrent latent state. SSVR encodes the initial image and instruction once, then conditions each action prediction on the latent state and updates it with an action-conditioned GRU. Using Qwen2.5-VL as the backbone, SSVR achieves 99.5/99.6, 96.3/98.0, and 83.9/90.6 EM/PR on FrozenLake, Maze, and MiniBehavior, substantially outperforming prior methods. Extensive experiments support the effectiveness of recurrent state modeling for VLA open-loop planning across input transformations and transfer settings.
By reusing static visual-textual context and updating a compact recurrent state, SSVR supports efficient multi-step inference, achieving up to $98.58\times$ faster Maze decoding rollouts than the evaluated baselines with the prefix cache prebuilt.
Code is publicly available at \href{https://anony-xyz.github.io/}{\texttt{project page}}.
\end{abstract}

\begin{figure}[t]
\centering
\includegraphics[width=.99\textwidth]{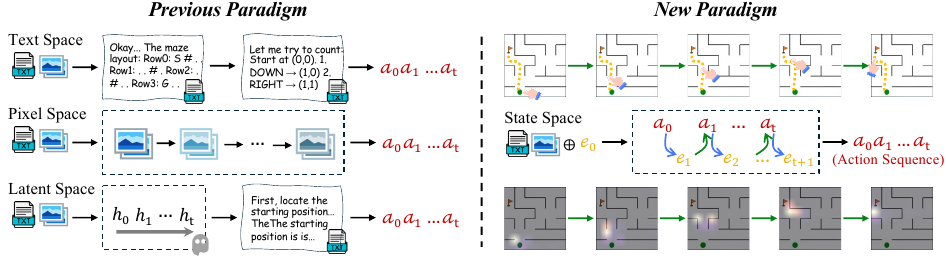}
\caption{Motivation and evidence for aligning reasoning with dynamic state. Existing paradigms represent evolving state through compressed language, repeatedly generated images, or intermediate latent tokens. SSVR follows the finger-tracing intuition: a static scene is reused, a recurrent latent state is updated by actions through a GRU, and visual attention varies across planning steps.}
\label{fig:motivation}
\end{figure}

\section{Introduction}

Vision-language-action (VLA) systems in robotics~\citep{look}, GUI interaction~\citep{iclrAgent} and navigation~\citep{pmlr-LM-Nav} often need to plan before execution because feedback may be delayed or costly and mistakes may be irreversible. In this open-loop setting, the model observes the initial scene and instruction once, internally models how actions change the execution state, and outputs a complete action sequence. Planning reliability therefore depends on maintaining an informative internal state throughout the rollout.

Existing reasoning paradigms exhibit a fundamental \emph{state-representation mismatch} with this requirement. They conflate two components with different temporal roles: high-dimensional scene context that remains largely static and execution state that evolves with actions. Text-space reasoning compresses map topology and dynamic state into language, making critical spatial relations easy to omit or blur; pixel-space reasoning repeatedly generates complete visual states, so lossy frame errors propagate into later decisions; and latent-space reasoning may bypass intermediate visual tokens. In open-loop rollout, textual omissions and pixel-generation errors can compound, while latent bypass can weaken the use of intermediate visual representations.

This asymmetry between static context and dynamic state is common in VLA open-loop planning~\citep{sokoban,frozenlake,maze,minibehav}. In the Maze example in Figure~\ref{fig:motivation}, the walls and goal remain fixed while only the agent position changes: prior paradigms repeatedly re-express the evolving state in text, images, or hidden vectors, whereas a human can trace the route by holding the map fixed and moving only a finger. Figure~\ref{fig:failures} shows the failures: topology hallucination in text-space reasoning, object duplication or disappearance in pixel-space rollout, and bypass of intermediate visual tokens in latent-space reasoning. Together, these observations motivate a reasoning paradigm better matched to open-loop state evolution.

We propose \textbf{State-Space Visual Reasoning} (SSVR), which addresses this mismatch by decoupling static visual context, language constraints, and a recurrent latent state while retaining the pretrained MLLM backbone~\citep{qwenvl}. The initial image and instruction are encoded once into a reusable KV cache~\citep{jiang2025purekv,jiang2026acckv}; at each step, the latent state token $e_t=z_t$ conditions next-action prediction, and the predicted action updates $z_t$ through a learned GRU transition, recursively advancing the latent state-action trajectory.

Experiments show three advantages of SSVR: \textbf{(1) strong planning performance}, with 99.5/99.6, 96.3/98.0, and 83.9/90.6 EM/PR on FrozenLake, Maze, and MiniBehavior, respectively, surpassing the strongest pixel baseline; \textbf{(2) robustness and generalization}, with resilience to prompt and image transformations, stronger FrozenLake-to-Maze transfer than VPRL~\citep{vp}, and a 22.9-point gain in native Maze VQA optimal-action hit rate without recurrent state conditioning (Table~\ref{tab:maze_vqa}), while VQAv2\footnote{\url{https://huggingface.co/datasets/HuggingFaceM4/VQAv2}}~\citep{vqav2} accuracy drops by only 3.9 points from native Qwen; and \textbf{(3) efficient reasoning}, with decoding-rollout speedups of $5.81\times$, $38.24\times$, and $98.58\times$ over LVR, Monet, and VPRL, respectively, using a prebuilt prefix cache.

\section{Related Work}

\noindent
\textbf{Text-space visual reasoning.} MLLMs often map image features into language-model space and use text tokens as intermediate reasoning carriers~\citep{llava, alayrac2022flamingo, liu2025can, venhoff2025visual}. CoT-style methods externalize spatial inference as symbols~\citep{wu2024mindseyellmsvisualizationofthought,ascii} or enrich text reasoning with visual abstraction, perception tokens, region distillation, and tool use~\citep{liu2025thinkingvisualabstractenhancing,bigverdi2024perceptiontokensenhancevisual,wei2026zoomingzoomingregiontoimagedistillation,zheng2026deepeyesincentivizingthinkingimages,DeepEyesV2}. This suits knowledge QA and compactly verbalizable vision tasks, but verbalizing map-like inputs is lossy: text must maintain coordinates, adjacency, reachability, and action history, making critical topology easy to omit or blur as the rollout evolves. SSVR keeps language for goals and constraints, while an action-conditioned GRU maintains the recurrent planning state.

\noindent
\textbf{Pixel-space visual reasoning.} Pixel-space methods construct explicit visual intermediates through drawing, annotation, programmatic editing, generated visual thoughts~\citep{li2025imaginereasoningspacemultimodal,chern2025thinkinggeneratedimages}, or future image/state rollouts for VLA and spatial planning~\citep{CoT-VLA,Wang2026VLAThinkerBV,vp}. They preserve spatial structure and improve inspectability, but open-loop tasks often change only low-dimensional variables, such as position, carrying state, or phase, while the scene stays mostly fixed. Full image rollout therefore regenerates mostly static content; more importantly, each generated frame is a lossy state estimate whose rendering errors can propagate and compound when reused by later steps. SSVR preserves visual grounding and decouples static context from recurrent state.

\noindent
\textbf{Latent-space visual reasoning.} Latent-space methods replace pixels with continuous tokens, reducing generation cost and enabling lightweight reasoning in a shared vision-language space~\citep{monet,lvr,liu2025reasoning,yang2025machinementalimageryempower,chen2026reasoning,tang2026thinking,atlas}. Prior analyses report image-to-latent and latent-to-answer disconnections~\citep{li2026imagination,latentsurvey}, alongside representation homogenization, manifold separation, and input-output embedding mismatch under recursive use~\citep{cui2026retrieveintegratesynthesizespatialsemantic,miao2026gapgranularalignmentparadigm}. SSVR uses a continuous recurrent state initialized from visual features, updated by an action-conditioned GRU, and injected as the query for action prediction.

\section{Vision-Language-Action Open-Loop Planning}
\label{sec:planning}

We study VLA open-loop planning. Given an initial image $I_0$ and instruction $x$, the model predicts an action sequence $\mathbf{a}=(a_0,\ldots,a_{H-1})$. The evaluation environment is defined as
\begin{equation}
\begin{array}{l}
\mathcal{T}=(I_0,x,\mathcal{A},s_0,T,g,H),\\
s_{t+1}=T(s_t,a_t),\quad a_t\in\mathcal{A},
\end{array}
\end{equation}
where $s_0$ is the initial environment state, $\mathcal{A}$ is the action set, $T$ is the environment transition, $g$ is the goal predicate, and $H$ is the planning horizon. The evaluator uses $T$ and $g$ to execute and score the plan; success requires $g(s_t)=1$ for some $t\le H$. In the default evaluation, $H$ is set to the ground-truth shortest-path length, and execution stops at the first terminal event.

Here, $I_0$ provides persistent scene context, $x$ specifies task objectives and constraints, and $s_t$ denotes the changing environment state. The model maintains a learned latent state $z_t$ from the initial visual features and action history. A suitable reasoning paradigm should reuse persistent context while updating its internal state according to predicted actions.

\begin{figure}[t]
\centering
\includegraphics[width=.99\textwidth]{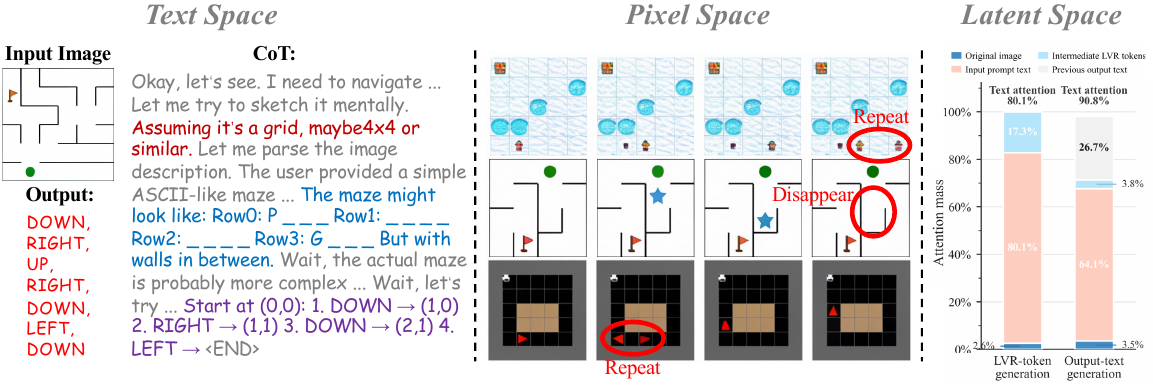}
\caption{Representative failure modes of existing reasoning paradigms. Text-space reasoning can hallucinate Maze topology after compressing the image into language; repeatedly generated pixel-space states may duplicate or erase objects and propagate these errors; latent-space reasoning can bypass both image and intermediate latent tokens when producing the final answer.}
\label{fig:failures}
\end{figure}

Figure~\ref{fig:failures} summarizes the resulting failures across existing paradigms, motivating a better reasoning paradigm.

\section{State-Space Visual Reasoning}
\label{sec:method}

\begin{figure}[t]
\centering
\includegraphics[width=0.9\textwidth]{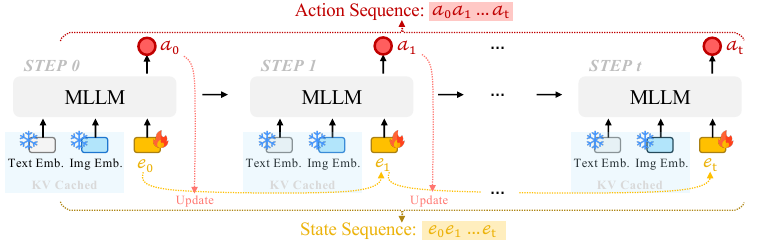}
\caption{Overview of SSVR. The initial image and instruction are encoded once and kept as cached visual-textual context. Visual features initialize the latent state $z_0$. At step $t$, the model appends the state token $e_t=z_t$, predicts action $a_t$, and updates the latent state through an GRU.}
\label{fig:ssvr}
\end{figure}

Figure~\ref{fig:ssvr} summarizes SSVR, which reuses static visual-textual context while recursively updating a GRU latent state.

\subsection{Static Context and Dynamic State}

Let $\theta=(\theta_0,\Delta\theta)$ collect the frozen pretrained parameters and trainable LoRA adapters, and let $\eta$ denote the trainable latent-state parameters. The decoder is $F_\theta$, and the visual frontend produces projected feature tokens $V_0=\mathcal{E}^{\mathrm{vis}}_\theta(I_0)\in\mathbb{R}^{N_v\times d}$. We pack these features with the instruction and template tokens into the input embedding sequence $C$ and compute the static prefix cache
\begin{equation}
(C,\mathcal{M})=\operatorname{Pack}_{\theta_0}(V_0,x),\qquad
\mathcal{K}=F_\theta^{\mathrm{prefill}}(C;\mathcal{M}).
\label{eq:static_context}
\end{equation}
The metadata $\mathcal{M}$ retains the attention mask and position information for decoding.

The GRU latent state $z_t\in\mathbb{R}^d$ represents the evolving planning state. Let $U_0\in\mathbb{R}^{N_u\times d}$ contain the initial scene-token embeddings used by the initializer; their construction is specified in Appendix~\ref{sec:app_initialization}. The initial state uses their mean:
\begin{equation}
c_0=\frac{1}{N_u}\sum_{p=1}^{N_u}U_{0,p},\qquad
z_0=f^{\mathrm{init}}_\eta(U_0)=\operatorname{LN}_\eta(q_\eta+W_\eta c_0),\qquad e_t=z_t,
\label{eq:gru_initialization}
\end{equation}
where $q_\eta$, $W_\eta$, and the LayerNorm parameters are learned jointly with the GRU transition.

At step $t$, SSVR appends $e_t$ as an independent sequence position after the cached context and computes the decoder readout
\begin{equation}
u_t=F_\theta^{\mathrm{decode}}(\mathcal{K},e_t;\mathcal{M}).
\label{eq:cached_decode}
\end{equation}
Each decision reuses the same prefix cache and appended-token position. The current state token's temporary KV entries are discarded after decoding, implementing the sequence $C\mathbin{\Vert}e_t$.

\subsection{State-Conditioned Action Prediction}

SSVR predicts one action at a time from the current latent state and cached context. Each $a\in\mathcal{A}$ is assigned a distinct single-token verbalizer, such as \texttt{UP}, \texttt{DOWN}, \texttt{LEFT}, or \texttt{RIGHT}. If $v(a)$ is its token ID, the action logit is
\begin{equation}
o_t(a)=\mathrm{LMHead}_{\theta_0}(u_t)_{v(a)},\quad a\in\mathcal{A},
\end{equation}
and the resulting policy is
\begin{equation}
\pi_{\theta,\eta}(a\mid C,z_t)=\frac{\exp(o_t(a))}{\sum_{a'\in\mathcal{A}}\exp(o_t(a'))}.
\label{eq:action_distribution}
\end{equation}
This reuses the pretrained language-model head, with probabilities normalized over the task action set.

\subsection{Recursive State-Space Inference}

With greedy decoding, SSVR alternates between state-conditioned action prediction and action-conditioned latent transition:
\begin{equation}
\begin{aligned}
a_t&=\arg\max_{a\in\mathcal{A}}\pi_{\theta,\eta}(a\mid C,z_t),\\
z_{t+1}&=f^{\mathrm{trans}}_\eta(z_t,a_t)=\operatorname{LN}_\eta\!\left(\operatorname{GRUCell}_\eta(E^a_\eta(a_t),z_t)\right),
\qquad e_{t+1}=z_{t+1}.
\end{aligned}
\label{eq:gru_transition}
\end{equation}
The learned action embedding $E^a_\eta(a_t)$ and previous latent state drive the GRU update. Initialization and transition share the LayerNorm module. This recursion produces an $H$-step plan using the fixed visual-textual context and predicted action history. The evaluator handles environment execution and terminal-event checks.

\subsection{Soft-Target Training}

For each training instance, environment transitions and path-quality rules identify safe actions $\mathcal{A}_{\mathrm{safe}}(s_t)$ and optimal actions $\mathcal{A}_{\mathrm{opt}}(s_t)$ offline at the reference decision state $s_t$. A reference action history is constructed by BFS from the episode start, and replayed through the GRU to obtain the supervised latent state $z_t$. For $\mathcal{A}_{\mathrm{opt}}(s_t)\subseteq\mathcal{A}_{\mathrm{safe}}(s_t)$ and a nonempty optimal set, the soft target is
\begin{equation}
q_\alpha(a\mid s_t)=(1-\alpha)\,\mathrm{Unif}(\mathcal{A}_{\mathrm{safe}}(s_t))(a)+\alpha\,\mathrm{Unif}(\mathcal{A}_{\mathrm{opt}}(s_t))(a).
\label{eq:soft_target}
\end{equation}
An empty optimal set uses the uniform safe-action target; training retains items with nonempty safe sets. SSVR minimizes
\begin{equation}
\mathcal{L}_{\mathrm{SFT}}=-\sum_{a\in\mathcal{A}}q_\alpha(a\mid s_t)\log\pi_{\theta,\eta}(a\mid C,z_t).
\label{eq:training_loss}
\end{equation}
Gradients propagate through the decoder, differentiable prefix cache, and reference-history GRU updates. Training jointly learns the LoRA adapters, latent initializer, action embeddings, and GRU transition while keeping pretrained weights frozen. The coefficient $\alpha$ balances safe-action diversity and optimal-action preference; the main model uses $\alpha=0.7$. Implementation details and pseudocode are provided in Appendix.

\section{Experiments}

We evaluate open-loop rollout planning on FrozenLake~\citep{frozenlake}, Maze~\citep{maze}, and MiniBehavior~\citep{minibehav}, covering risk avoidance, topological path finding, and phased object interaction, respectively. The model uses the initial observation and instruction, and updates its latent state with its own predicted actions. Metrics are exact match (EM) and progress rate (PR)~\citep{vp}, evaluated against shortest optimal action sequences; full implementation, training, and evaluation settings are in the appendix.

\subsection{Main Results}


\begin{table}[t]
\centering
\caption{Comparison across reasoning paradigms. Entries report EM/PR (\%). VQA Support: native VQA through text-based responses; $^*$: official release without training on our datasets.}
\label{tab:main}
\resizebox{\textwidth}{!}{
\begin{tabular}{lccccc}
\hline
Model & VQA Support & FrozenLake & Maze & MiniBehavior & Avg. \\
\hline
\multicolumn{6}{c}{\cellcolor[HTML]{F2F2F2}\textit{Text Space}} \\
Gemini 2.0 Flash Direct$^*$ & \CheckmarkBold & 21.2 / 47.6 & 8.3 / 31.4 & 0.7 / 29.8 & 10.1 / 36.3 \\
Gemini 2.0 Flash CoT$^*$ & \CheckmarkBold & 27.6 / 52.5 & 6.9 / 29.8 & 4.0 / 31.2 & 12.8 / 37.8 \\
Gemini 2.5 Pro$^*$ & \CheckmarkBold & 72.0 / 85.0 & 21.5 / 35.5 & 37.6 / 59.9 & 43.7 / 60.1 \\
Qwen2.5-VL-7B Direct$^*$ & \CheckmarkBold & 1.2 / 15.0 & 0.6 / 14.5 & 0.3 / 9.8 & 0.7 / 13.1 \\
Qwen2.5-VL-7B CoT$^*$ & \CheckmarkBold & 8.2 / 29.1 & 2.3 / 15.2 & 0.5 / 14.7 & 3.7 / 19.7 \\
Qwen2.5-VL-7B SFT & \CheckmarkBold & 68.6 / 84.4 & 60.9 / 70.3 & 31.3 / 56.1 & 53.6 / 70.3 \\
\multicolumn{6}{c}{\cellcolor[HTML]{F2F2F2}\textit{Pixel Space}} \\
VPFT & \XSolidBrush & 75.4 / 79.5 & 59.0 / 64.0 & 33.8 / 52.2 & 56.1 / 65.2 \\
VPRL & \XSolidBrush & \underline{91.6} / \underline{93.2} & \underline{74.5} / \underline{77.6} & \underline{75.8} / \underline{83.8} & \underline{80.6} / \underline{84.9} \\
\multicolumn{6}{c}{\cellcolor[HTML]{F2F2F2}\textit{Latent Space}} \\
LVR$^*$ & \CheckmarkBold & 6.8 / 16.0 & 5.0 / 10.2 & 0.0 / 8.0 & 3.9 / 11.4 \\
Monet$^*$ & \CheckmarkBold & 5.0 / 13.0 & 2.3 / 5.0 & 0.0 / 7.5 & 2.4 / 8.5 \\
\multicolumn{6}{c}{\cellcolor[HTML]{F2F2F2}\textit{Recurrent State (Ours)}} \\
SSVR & \CheckmarkBold & \textbf{99.5} / \textbf{99.6} & \textbf{96.3} / \textbf{98.0} & \textbf{83.9} / \textbf{90.6} & \textbf{93.2} / \textbf{96.1} \\
\hline
\end{tabular}
}
\end{table}

Table~\ref{tab:main} reports aggregate performance. SSVR achieves the highest EM and PR on all three tasks, averaging 93.2/96.1 and outperforming the strongest pixel baseline, VPRL, by 12.6/11.2 percentage points. The largest improvement occurs on Maze, where EM/PR rises from 74.5/77.6 to 96.3/98.0, a gain of 21.8/20.4 points. SSVR also achieves 99.5/99.6 on FrozenLake and 83.9/90.6 on MiniBehavior, compared with VPRL's 91.6/93.2 and 75.8/83.8, respectively. These gains are consistent with the benefit of conditioning decisions on an updated recurrent state, although MiniBehavior remains more challenging than the navigation tasks.

The baseline pattern is consistent with the proposed \emph{state-representation mismatch}. In text space, verbalizing a map is a lossy compression: walls, narrow passages, and adjacency relations are easily blurred, and these omissions enter the evolving textual state. Task-specific text SFT reaches 53.6/70.3 EM/PR. In pixel space, each generated frame can introduce errors, including duplicated or disappearing objects in Figure~\ref{fig:failures}. Subsequent steps condition on these generated states, allowing errors to propagate with rollout length. The evaluated latent-space baselines exhibit bypass and weak transfer. On our tasks, Monet ceases to emit latent tokens and falls back to text-only reasoning. During output generation, LVR assigns 3.5\% attention to image tokens and 3.8\% to intermediate latent tokens, while text tokens dominate (Figure~\ref{fig:failures}), indicating limited direct attention to visual and intermediate latent representations. The released LVR and Monet models average 3.9\% and 2.4\% EM, respectively, in this transfer setting. SSVR learns an action-conditioned recurrent update and uses the resulting latent state for each action decision.

\begin{figure}[t]
\centering
\begin{minipage}[t]{.75\textwidth}
\centering
\includegraphics[width=\linewidth]{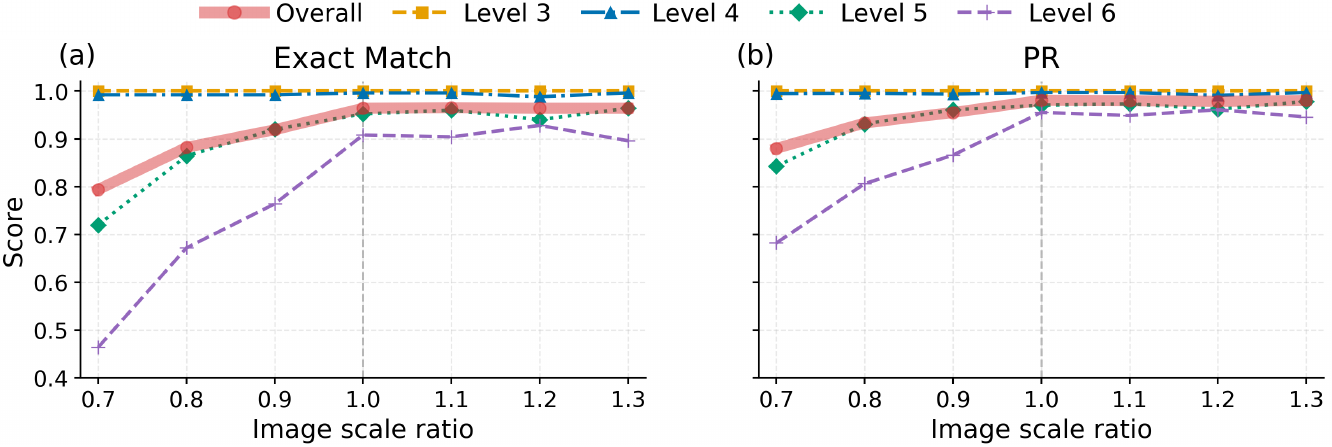}
\caption{Maze EM and PR across image scale ratios 0.7--1.3, overall and by difficulty level.}
\label{fig:ratio}
\end{minipage}
\end{figure}

\subsection{Robustness and Generalization}

We evaluate SSVR across prompt paraphrasing, image transformations, cross-dataset transfer, and VQA retention.

\begin{table}[t]
\centering
\begin{minipage}[t]{.52\textwidth}
\centering
\caption{Prompt paraphrase robustness (\%).}
\label{tab:prompt}
\small
\setlength{\tabcolsep}{4pt}
\begin{tabular}{lrrr}
\toprule
Prompt & EM & PR & Legal \\
\midrule
Default & \underline{96.3} & \underline{98.0} & \textbf{100.0} \\
Paraphrased & \textbf{96.8} & \textbf{98.3} & \textbf{100.0} \\
\bottomrule
\end{tabular}
\end{minipage}\hfill
\begin{minipage}[t]{.45\textwidth}
\centering
\caption{Native VQAv2 retention.}
\label{tab:vqa}
\small
\setlength{\tabcolsep}{4pt}
\begin{tabular}{lrr}
\toprule
Model & Acc. (\%) & $\Delta$ (pp) \\
\midrule
Native Qwen & \textbf{71.9} & 0.0 \\
SSVR & \underline{68.0} & $-3.9$ \\
VPRL & \XSolidBrush & \XSolidBrush \\
\bottomrule
\end{tabular}
\end{minipage}
\end{table}

\subsubsection{Input Robustness}

With semantically equivalent prompt paraphrases (Table~\ref{tab:prompt}), EM increases from 96.3 to 96.8 and PR from 98.0 to 98.3, while the legal rate remains 100\%. These small changes indicate stable performance under semantically equivalent wording.

We resize Maze inputs over ratios 0.7--1.3 to test scale robustness. Figure~\ref{fig:ratio} reveals a difficulty-dependent response: Level 3 maintains 100\% EM/PR throughout, and Level 4 EM remains between 98.8\% and 99.6\%. Shrinking substantially affects harder maps: at scale 0.7, Level-5 EM/PR falls from 95.20/97.03 to 72.00/84.26, and Level-6 EM/PR from 90.80/95.52 to 46.40/68.22, relative to scale 1.0. Performance under enlargement remains close to the native-scale baseline; at scale 1.3, Level-5 and Level-6 EM/PR are 96.40/97.81 and 89.60/94.56, respectively. This asymmetry suggests sensitivity to the loss of fine visual detail on difficult maps under downscaling, together with stable performance under enlargement.

We further invert colors and synchronously flip images, evaluator coordinates, and action labels to test sensitivity to appearance and orientation. Table~\ref{tab:transform} shows that color inversion reduces EM from 96.30 to 94.00 and PR from 97.98 to 96.57, with the legal rate remaining at 99.90\%. Vertical and horizontal flips each reduce EM by 0.30 points to 96.00, while flipping both directions reduces it by 1.30 points to 95.00; all flips retain a 100\% legal rate. These modest declines indicate robustness to the tested appearance and orientation changes, with greater sensitivity to color inversion. Since these transformations preserve topology, the results are consistent with the use of relational geometry.

\begin{table}[t]
\centering
\begin{minipage}[t]{.54\textwidth}
\centering
\caption{Color/flip robustness.}
\label{tab:transform}
\setlength{\tabcolsep}{3pt}
\footnotesize
\begin{tabular}{lrrr}
\toprule
Transformation & EM & PR & Legal Rate \\
\midrule
Original & 96.30 & 97.98 & 100.00 \\
Color inversion & 94.00$^{-2.30}$ & 96.57$^{-1.41}$ & 99.90$^{-0.10}$ \\
Vertical flip & 96.00$^{-0.30}$ & 97.62$^{-0.36}$ & 100.00$^{0.00}$ \\
Horizontal flip & 96.00$^{-0.30}$ & 97.77$^{-0.21}$ & 100.00$^{0.00}$ \\
Dual-direction flip & 95.00$^{-1.30}$ & 97.00$^{-0.98}$ & 100.00$^{0.00}$ \\
\bottomrule
\end{tabular}
\end{minipage}\hfill
\begin{minipage}[t]{.43\textwidth}
\centering
\caption{FrozenLake$\rightarrow$Maze transfer. BB+MazeIF uses a Maze-trained interface.}
\label{tab:ood}
\setlength{\tabcolsep}{3pt}
\footnotesize
\begin{tabular}{lrr}
\hline
Method & Maze EM & Maze PR \\
\hline
Qwen Direct & 0.6 & 14.5 \\
Qwen CoT & 2.3 & 15.2 \\
VPRL & 6.8 & 7.2 \\
SSVR (BB+MazeIF) & \underline{30.9} & \underline{36.9} \\
SSVR (Full) & \textbf{33.6} & \textbf{38.3} \\
\hline
\end{tabular}
\end{minipage}
\end{table}

\subsubsection{Cross-Dataset Transfer}

\begin{table}[t]
\centering
\setlength{\tabcolsep}{4pt}
\caption{Native Maze action-choice VQA in generation mode.}
\label{tab:maze_vqa}
\small
\begin{tabular}{c c c c c}
\toprule
Level & Base Opt. & Trained Opt. & Base Legal & Trained Legal \\
\midrule
3 & 24.00 & 50.40$^{+26.40}$ & 43.60 & 72.80$^{+29.20}$ \\
4 & 22.80 & 49.20$^{+26.40}$ & 49.60 & 72.80$^{+23.20}$ \\
5 & 24.00 & 44.80$^{+20.80}$ & 53.60 & 72.80$^{+19.20}$ \\
6 & 24.40 & 42.40$^{+18.00}$ & 60.00 & 73.60$^{+13.60}$ \\
\midrule
Avg. & \textbf{23.80} & \textbf{46.70}$^{+22.90}$
& \textbf{51.70} & \textbf{73.00}$^{+21.30}$ \\
\bottomrule
\end{tabular}
\end{table}

FrozenLake$\rightarrow$Maze transfer shifts from avoiding circular holes to parsing linear wall topology. It changes both visual primitives and connectivity cues, testing transferable path finding. \texttt{Full} applies the FrozenLake-trained backbone and GRU module directly to Maze. \texttt{BB+MazeIF} combines the FrozenLake-trained backbone adapters with the Maze-trained GRU/action interface, evaluating compatibility across separately trained components.

Table~\ref{tab:ood} shows that both variants substantially outperform Qwen and VPRL. \texttt{Full} raises EM from 2.3\% for Qwen-CoT and 6.8\% for VPRL to 33.6\%. This transfer from holes to wall topology is consistent with learned use of connectivity cues. The gains also persist on hard maps: \texttt{Full} reaches 27.2/32.1 EM/PR on Level 5 and 22.4/29.9 on Level 6, versus VPRL's 2.4/2.6 and 4.0/4.6. \texttt{BB+MazeIF} is slightly weaker than \texttt{Full}, possibly because replacing the GRU interface disrupts the joint calibration learned during training.

\subsubsection{VQA Retention}

We disable the GRU state-conditioning pathway and evaluate the trained backbone through its original LM head on native Maze action-choice VQA and VQAv2, using an LLM judge for VQAv2 semantic correctness. Table~\ref{tab:maze_vqa} shows that average optimal-action and legal-action hit rates rise from 23.8\% to 46.7\% and from 51.7\% to 73.0\%, respectively, with gains at every difficulty level. On VQAv2 (Table~\ref{tab:vqa}), SSVR retains 68.0\% accuracy versus native Qwen's 71.9\%, a 3.9-point decrease; VPRL does not support native text-response VQA. Thus, training with the GRU state module improves the backbone's native spatial action-choice capability while largely preserving general VQA ability.

\subsection{Ablations}

We analyze two key designs: soft supervision and the action-output interface.

\subsubsection{Soft-Target Coefficient}

\begin{table}[t]
\centering
\caption{Soft-target coefficient ablation on Maze.}
\label{tab:alpha}
\resizebox{\textwidth}{!}{
\begin{tabular}{lrrrrrrr}
\hline
$\alpha$ & Maze EM & Maze PR & Entropy & VQAv2 Acc. & Paraphrase EM & Scale 0.7 EM & Scale 1.3 EM \\
\hline
1.0 & 91.8 & 95.7 & 0.0045 & 67.6 & 91.7 & 68.7 & 90.5 \\
0.7 & \textbf{96.3} & \textbf{98.0} & 0.4833 & \underline{68.0} & \textbf{96.8} & \underline{79.4} & \textbf{96.4} \\
0.4 & \underline{95.5} & \underline{97.5} & 0.7287 & \textbf{71.1} & \underline{94.7} & \textbf{80.0} & \underline{95.1} \\
\hline
\end{tabular}
}
\end{table}

We vary only $\alpha$ to examine the balance between safe-action and optimal-action supervision. Table~\ref{tab:alpha} shows that reducing $\alpha$ from 1.0 to 0.7 increases action entropy from 0.0045 to 0.4833 and improves Maze EM/PR from 91.8/95.7 to 96.3/98.0, with gains in all other evaluated metrics. Further reducing $\alpha$ to 0.4 raises entropy to 0.7287 and VQAv2 accuracy from 68.0 to 71.1, while slightly improving EM at scale 0.7 from 79.4 to 80.0. Maze EM/PR falls to 95.5/97.5, paraphrase EM drops from 96.8 to 94.7, and EM at scale 1.3 decreases from 96.4 to 95.1. Greater safe-action weight thus yields higher policy entropy with differing effects across planning and robustness metrics. We use $\alpha=0.7$ because it achieves the best in-domain EM/PR, paraphrase EM, and enlarged-image EM among the tested settings, while $\alpha=0.4$ favors VQA retention and robustness to image shrinking.

\subsubsection{Action Output}

\begin{table}[t]
\centering
\caption{Action-output ablation.}
\label{tab:state_head}
\small
\begin{tabular}{lrrr}
\toprule
Variant & Maze EM (\%) & Maze PR (\%) & VQAv2 (\%) \\
\midrule
SSVR & \underline{96.3} & \underline{98.0} & \textbf{68.0} \\
SSVR-head & \textbf{96.4}$^{+0.1}$ & \textbf{98.5}$^{+0.5}$ & \underline{46.9}$^{-21.1}$ \\
\bottomrule
\end{tabular}
\end{table}

Table~\ref{tab:state_head} shows that replacing the LM action head with a linear softmax classifier, while keeping the GRU fixed, increases Maze EM/PR from 96.3/98.0 to 96.4/98.5, gains of 0.1/0.5 percentage points, and reduces VQAv2 accuracy from 68.0 to 46.9, a loss of 21.1 points. VQAv2 uses native text generation through the original LM head for both variants. Reusing the LM head during planning training therefore retains more general VQA capability in this comparison while achieving comparable planning performance.

\section{Discussion}

\subsection{Interpretability Analysis}

\begin{figure}[t]
\centering
\includegraphics[width=.8\textwidth]{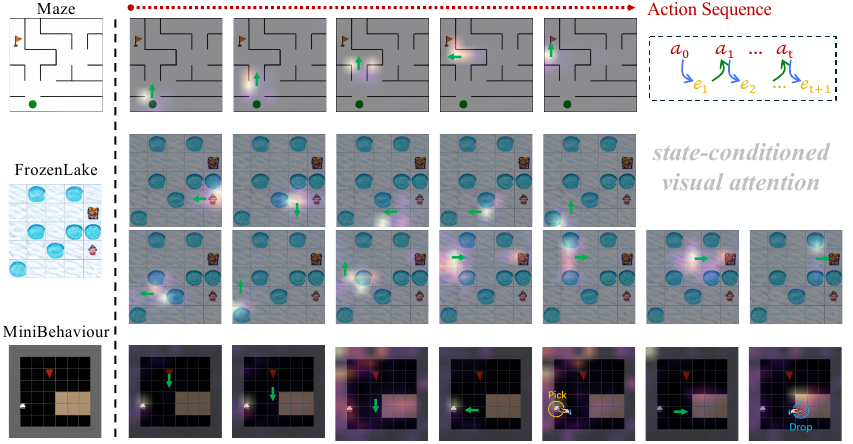}
\caption{Gradient-weighted visual attribution from the GRU state-token query to image tokens. As the recurrent latent state changes, attribution highlights next-action-relevant regions: nearby walls in Maze, nearby holes in FrozenLake, and a phase-dependent shift from printer to table in MiniBehavior.}
\label{fig:attention}
\end{figure}

We use gradient-weighted attention to visualize attribution from the GRU state-token query to image tokens, conditioned on the selected action logit. Figure~\ref{fig:attention} shows changing attribution across planning steps: nearby walls in Maze, nearby holes in FrozenLake, and a shift from printer to table across the \texttt{PICK}/\texttt{DROP} phases in MiniBehavior. This analysis measures the current decision's visual readout with the latent state and prefix cache held fixed, illustrating how static context is read under the evolving recurrent state.

\subsection{Efficiency Analysis}

\subsubsection{Qualitative Analysis}

SSVR's efficiency advantage comes from static-context reuse and compact recurrent updates. Let $N_C$ be the number of static-context tokens, $N_E=1$ the state-token count, $N_I$ the number of intermediate visual-frame tokens, and $H$ the planning horizon. With fixed decoder width and depth, the attention cost with a static prefix KV cache is
\begin{equation}
\mathcal{O}\left(N_C^2 + H(N_EN_C+N_E^2)\right).
\end{equation}
The GRU additionally costs $\mathcal{O}(Hd^2)$ for input and hidden dimension $d$. A visual planner that processes one $N_I$-token intermediate frame per step against a fixed prefix has attention cost
\begin{equation}
\mathcal{O}\left(N_C^2 + H(N_IN_C+N_I^2)\right).
\end{equation}
These expressions describe attention over the stated token sequences; total runtime also includes projections, feed-forward layers, and visual processing. Since $N_E\ll N_I$, SSVR reduces the per-step token workload while reusing the visual-textual prefix.

\subsubsection{Quantitative Analysis}

\begin{table}[t]
\centering
\caption{Efficiency on Maze \texttt{case0}. $^\dagger$: KV cache; latency excludes preprocessing and, for SSVR$^\dagger$, prefix-cache construction. Multipliers are relative to SSVR$^\dagger$.}
\label{tab:efficiency}
\resizebox{.85\textwidth}{!}{
\begin{tabular}{lrrrr}
\hline
Method & Step latency (ms) & Rollout latency (ms) & Throughput (cases/s) & Peak memory (GiB) \\
\hline
SSVR$^\dagger$ & \textbf{45.611} {\scriptsize(1.00x)} & \textbf{182.442} {\scriptsize(1.00x)} & \textbf{5.481} {\scriptsize(1.00x)} & 15.918 {\scriptsize(1.00x)} \\
SSVR & \underline{89.159} {\scriptsize(1.95x)} & \underline{356.634} {\scriptsize(1.95x)} & \underline{2.804} {\scriptsize(0.51x)} & 15.937 {\scriptsize(1.00x)} \\
LVR & 529.596 {\scriptsize(11.61x)} & 1059.193 {\scriptsize(5.81x)} & 0.944 {\scriptsize(0.17x)} & 17.609 {\scriptsize(1.11x)} \\
Monet & 1743.966 {\scriptsize(38.24x)} & 6975.866 {\scriptsize(38.24x)} & 0.143 {\scriptsize(0.03x)} & \underline{15.705} {\scriptsize(0.99x)} \\
VPRL & 4496.111 {\scriptsize(98.58x)} & 17984.442 {\scriptsize(98.58x)} & 0.056 {\scriptsize(0.01x)} & \textbf{13.331} {\scriptsize(0.84x)} \\
\hline
\end{tabular}
}
\end{table}

We benchmark all methods on the same Maze \texttt{case0} sample over 20 repetitions after warm-up, using CUDA events for latency and the PyTorch CUDA memory API for peak memory. Image/text preprocessing and prefix-cache construction are measured separately; cached SSVR rollout timing covers decoding with the prebuilt cache and recurrent state updates.

As shown in Table~\ref{tab:efficiency}, enabling KV cache reduces SSVR step latency from 89.159 ms to 45.611 ms and decoding-rollout latency from 356.634 ms to 182.442 ms, giving about a $1.95\times$ speedup with nearly unchanged peak memory. Compared with LVR, Monet, and VPRL, cached SSVR has $5.81\times$, $38.24\times$, and $98.58\times$ lower rollout latency under this timing protocol, illustrating the computational benefit of reusing static visual context during recurrent planning.

\section{Conclusion}

This paper identifies \emph{state-representation mismatch} as an obstacle in VLA open-loop planning: existing paradigms can encode evolving state through lossy textual compression, error-accumulating pixel rollouts, or bypass-prone latent tokens, even though most visual context remains static. SSVR addresses this mismatch by encoding the initial visual-textual context once and recursively updating a learned latent state through an action-conditioned GRU. Across three planning tasks, this design yields strong in-domain performance, robust behavior under linguistic and visual transformations, cross-dataset transfer, and efficient cached decoding while largely retaining general VQA ability. These results support action-conditioned recurrent state modeling as a useful design principle for VLA open-loop planning. 

\clearpage
\section*{AI-Assisted Content Disclosure}

Generative AI tools assisted with linguistic polishing and technical revision of the manuscript, including method descriptions and mathematical notation. The authors are responsible for verifying the revised text against the implementation and experimental records and for all claims, content, and findings in the paper.

\bibliographystyle{iclr2027_conference}
\bibliography{ssvr_iclr2027}
\clearpage
\appendix
\input{appendix/app}

\end{document}

%% file: appendix/app.tex
\newenvironment{apprevision}{\begingroup\color{black}}{\par\endgroup}

\section{Evaluation Metrics}

We use Exact Match (EM) and Progress Rate (PR) to evaluate open-loop plans. Consider an instance with $M$ shortest optimal action sequences of length $n>0$,
\begin{equation}
\mathbf{a}^{(m)}=(a_1^{(m)},\ldots,a_n^{(m)}),
\qquad m\in\{1,\ldots,M\},
\end{equation}
where $a_j^{(m)}$ is the $j$-th action of optimal sequence $m$. Let the predicted action sequence under the evaluation budget $H=n$ be
$\hat{\mathbf{a}}=(\hat a_1,\ldots,\hat a_n)$;
missing actions are treated as mismatches. The instance-level EM is
\begin{equation}
\mathrm{EM}=
\max_{m\in\{1,\ldots,M\}}
\prod_{j=1}^{n}\mathbb{I}\!\left[\hat a_j=a_j^{(m)}\right].
\label{eq:app_em}
\end{equation}
Thus, EM is one if the prediction exactly follows any shortest optimal action sequence and zero otherwise. The maximum accounts for all equally short optimal action sequences.

PR measures the longest optimal prefix:
\begin{equation}
\mathrm{PR}=
\max_{m\in\{1,\ldots,M\}}
\frac{1}{n}\sum_{j=1}^{n}
\prod_{k=1}^{j}\mathbb{I}\!\left[\hat a_k=a_k^{(m)}\right].
\label{eq:app_pr}
\end{equation}
If the first $\ell$ actions form an optimal prefix and the first error occurs at action $\ell+1$, then $\mathrm{PR}=\ell/n$. In the deterministic, unit-cost evaluation environments, an action is optimal when its transition is valid and reduces the task-specific shortest-path distance-to-goal by exactly one; successful termination has distance zero. This distance-based test identifies prefixes of shortest optimal action sequences. Dataset-level EM and PR are the arithmetic means of the corresponding instance-level scores and are reported as percentages in the main paper.

\section{Training and Inference Algorithms}
\label{sec:app_algorithms}

Algorithms~\ref{alg:ssvr_training} and~\ref{alg:ssvr_inference} implement the method in Section~\ref{sec:method}, using $\theta=(\theta_0,\Delta\theta)$ and latent-state parameters $\eta$. The initializer $f^{\mathrm{init}}_\eta$ and transition $f^{\mathrm{trans}}_\eta$ follow Equations~\ref{eq:gru_initialization} and~\ref{eq:gru_transition}.

\subsection{Implementation Details}

\paragraph{Packing and batching.}
The metadata $\mathcal{M}$ contains the prefix attention mask, multimodal position indices, appended-token position, and image-token key indices. Initializer pooling includes valid scene-token embeddings $U_0$ as defined in Section~\ref{sec:app_initialization}. Padded reference actions preserve the latent state during batched history replay.

\paragraph{Cache handling.}
An in-place cache is restored to the prefix length $N_C$ after each decode, preserving tensors required for training gradients. Training computes fresh differentiable caches for each minibatch. A full-sequence forward on $C\mathbin{\Vert}e_t$ provides an equivalent formulation in evaluation mode, where dropout is disabled.

\paragraph{Action-head ablation.}
The classifier variant uses $(W_\omega u_t+b_\omega)_a$ for the action logit and includes $\omega$ in the optimizer and saved model. The GRU update and soft-target supervision follow the main method.

\subsection{Training Pseudocode}

The training set $\mathcal{D}$ contains tuples $(I_0,x,\mathbf{a}^{\mathrm{ref}}_{<t},\mathcal{A}_{\mathrm{safe}},\mathcal{A}_{\mathrm{opt}})$. Each tuple supervises one reachable decision state; an empty history denotes the initial decision. The safe set is the base admissible set after the supervision protocol's action filters. Algorithm~\ref{alg:ssvr_training} averages the per-item loss in Equation~\ref{eq:training_loss} over the minibatch. In the pseudocode, $q_\alpha(a)$ denotes the target at the reference decision state of the current item.

\begin{algorithm}[!htbp]
\caption{Soft-target training with a GRU latent state}
\label{alg:ssvr_training}
\begin{algorithmic}[1]
\Require $\mathcal{D}$ as defined above; frozen pretrained parameters $\theta_0$; trainable LoRA parameters $\Delta\theta$; latent initializer $f^{\mathrm{init}}_\eta$; transition $f^{\mathrm{trans}}_\eta$; injective action verbalizer $v$; $\alpha\in[0,1]$
\Ensure Trained $\Delta\theta$ and $\eta$
\State Initialize $\Delta\theta$ and $\eta$; keep $\theta_0$ frozen
\For{each training minibatch $\mathcal{B}\subset\mathcal{D}$}
    \State Clear optimizer gradients; $\mathcal{L}\gets0$
    \For{each $(I_0,x,\mathbf{a}^{\mathrm{ref}}_{<t},\mathcal{A}_{\mathrm{safe}},\mathcal{A}_{\mathrm{opt}})\in\mathcal{B}$}
        \State $V_0\gets\mathcal{E}^{\mathrm{vis}}_\theta(I_0)$
        \State $(C,\mathcal{M})\gets\operatorname{Pack}_{\theta_0}(V_0,x)$
        \State $\mathcal{K}\gets F_\theta^{\mathrm{prefill}}(C;\mathcal{M})$
        \Comment{retain the training computation graph}
        \State Construct $U_0$ from the initial scene; $z\gets f^{\mathrm{init}}_\eta(U_0)$
        \For{$k=0,\ldots,t-1$}
            \State $z\gets f^{\mathrm{trans}}_\eta(z,a^{\mathrm{ref}}_k)$
            \Comment{differentiable reference-history replay}
        \EndFor
        \State $e\gets z$
        \State $u\gets F_\theta^{\mathrm{decode}}(\mathcal{K},e;\mathcal{M})$
        \Comment{read-only prefix cache; one appended token}
        \For{each action $a\in\mathcal{A}$}
            \State $o(a)\gets\operatorname{LMHead}_{\theta_0}(u)_{v(a)}$
        \EndFor
        \State $\pi_{\theta,\eta}(a\mid C,z)\gets\operatorname{softmax}_{a\in\mathcal{A}}(o(a))$
        \State Construct $q_\alpha$ using Equation~\ref{eq:soft_target} and its empty-optimal-set fallback
        \State $\mathcal{L}\gets\mathcal{L}-\sum_{a\in\mathcal{A}}q_\alpha(a)\log\pi_{\theta,\eta}(a\mid C,z)$
    \EndFor
    \State Backpropagate $\mathcal{L}/|\mathcal{B}|$ through decoding, prefill, and all replayed GRU transitions
    \State Update only $(\Delta\theta,\eta)$ using the optimizer
    \State Discard all minibatch KV caches and recurrent-state computation graphs
\EndFor
\State \Return $\Delta\theta,\eta$
\end{algorithmic}
\end{algorithm}

\subsection{Inference Pseudocode}

Algorithm~\ref{alg:ssvr_inference} uses the horizon and evaluation protocol in Section~\ref{sec:planning}. Greedy selection uses a fixed tie-breaking rule. Every predicted action drives the GRU transition, including actions marked invalid by the evaluator.

\begin{algorithm}[!htbp]
\caption{Open-loop GRU rollout with a read-only visual--textual KV cache}
\label{alg:ssvr_inference}
\begin{algorithmic}[1]
\Require Initial image $I_0$; instruction $x$; planning horizon $H$; trained parameters $(\theta_0,\Delta\theta,\eta)$; action verbalizer $v$
\Ensure Action plan $\mathbf{a}$ of length $H$
\State Set all modules to evaluation mode; disable gradients for ordinary rollout
\State $V_0\gets\mathcal{E}^{\mathrm{vis}}_\theta(I_0)$
\State $(C,\mathcal{M})\gets\operatorname{Pack}_{\theta_0}(V_0,x)$
\State $\mathcal{K}\gets F_\theta^{\mathrm{prefill}}(C;\mathcal{M})$
\Comment{compute the static prefix once}
\State Construct $U_0$ from the initial scene; $z\gets f^{\mathrm{init}}_\eta(U_0)$; $\mathbf{a}\gets[\ ]$
\For{$t=0,\ldots,H-1$}
    \State $e_t\gets z$
    \State $u_t\gets F_\theta^{\mathrm{decode}}(\mathcal{K},e_t;\mathcal{M})$
    \Comment{same prefix and appended position; discard temporary state KV}
    \For{each action $a\in\mathcal{A}$}
        \State $o_t(a)\gets\operatorname{LMHead}_{\theta_0}(u_t)_{v(a)}$
    \EndFor
    \State $a_t\gets\arg\max_{a\in\mathcal{A}}o_t(a)$
    \Comment{use a fixed tie-breaking rule}
    \State Append $a_t$ to $\mathbf{a}$
    \State $z\gets f^{\mathrm{trans}}_\eta(z,a_t)$
    \Comment{learned action-conditioned latent update}
\EndFor
\State \Return $\mathbf{a}$
\end{algorithmic}
\end{algorithm}

\section{State-Conditioned Visual Attribution Analysis}

This section defines visual attribution for the GRU latent-state variant under the same projected-image-token and cached-decoding assumptions. At step $t$, the conditioning state is the recurrent vector $z_t$ derived from the initial image features and the predicted action history.

\subsection{Gradient-Weighted Attribution Method}

\paragraph{Attention indexing under cached decoding.}
Let $\mathcal{P}\subset\{0,\ldots,N_C-1\}$ be the image-token \emph{key positions} in the packed static prefix. During a one-token cached decode, each decoder layer has one current query and $N_C+1$ keys, including the current state token. For decoder layer $\ell$ and query-attention head $j$, write the relevant attention matrix as
\begin{equation}
A_t^{\ell,j}\in\mathbb{R}^{1\times(N_C+1)}.
\end{equation}
The state query has local row index $0$ and full-sequence position $N_C$. The state-to-image attention is
\begin{equation}
a_{t,p}^{\ell,j}=A_t^{\ell,j}[0,p],
\qquad p\in\mathcal{P}.
\label{eq:app_raw_attention}
\end{equation}
For a full-sequence forward on $C\mathbin{\Vert}e_t$, the query row is $N_C$. Image-token indices follow the packed key order, including interleaved special and text tokens. In grouped-query attention, the averaging below uses query-attention heads.

\paragraph{Diagnostic forward with gradients.}
Ordinary rollout uses inference without gradients. For attribution, we snapshot the pre-action latent state $z_t$, retain the same static prefix and selected action, and perform a separate diagnostic one-token decode with gradient tracking enabled. The diagnostic pass holds model parameters and the stored rollout state fixed, with dropout disabled. The state snapshot and prefix KV cache are detached from the rollout graph, and a fresh state-embedding leaf enables gradient tracking for the diagnostic decode.

The diagnostic forward exposes the differentiable attention probabilities used in the attention-value product. An explicit attention implementation supports this gradient calculation. Numerical agreement with the ordinary decoder is checked when switching attention implementations, and the explained action remains fixed to the rollout selection. Temporary state-token KV entries are discarded after decoding.

\paragraph{Action-conditioned gradient.}
Let the action selected by the ordinary rollout be
\begin{equation}
\hat a_t=\arg\max_{a\in\mathcal{A}}o_t(a).
\end{equation}
The selected action $\hat a_t$ is held fixed during differentiation. Let $o_t^{\mathrm{diag}}(\hat a_t)$ denote the corresponding pre-softmax action logit in the diagnostic forward. Its derivative with respect to the attention tensor is
\begin{equation}
g_{t,p}^{\ell,j}
=\left[
  \frac{\partial o_t^{\mathrm{diag}}(\hat a_t)}
       {\partial A_t^{\ell,j}}
 \right]_{0,p},
\qquad p\in\mathcal{P}.
\label{eq:app_attention_gradient}
\end{equation}
Computationally, the gradient is obtained with respect to the attention tensor used by the forward pass and is then indexed at $[0,p]$.

We retain positive gradient-weighted attention and average it over the selected decoder layers $\mathcal{L}$ and their query-attention heads:
\begin{equation}
r_t(p)=\frac{1}{|\mathcal{L}|}
\sum_{\ell\in\mathcal{L}}\frac{1}{N_h^{\ell}}
\sum_{j=1}^{N_h^{\ell}}
\left[a_{t,p}^{\ell,j}g_{t,p}^{\ell,j}\right]_+,
\qquad [z]_+=\max(z,0).
\label{eq:app_attention_score}
\end{equation}
Here $N_h^{\ell}$ is the number of query-attention heads at layer $\ell$; if it is constant across layers, this reduces to division by $|\mathcal{L}|N_h$. A positive product indicates positive local sensitivity of the selected action logit to the corresponding attention entry, weighted by the attention magnitude.

For visualization, the scores are normalized separately at each step:
\begin{equation}
\widetilde r_t(p)=
\frac{r_t(p)}{\max_{p'\in\mathcal{P}}r_t(p')+\epsilon},
\qquad \epsilon>0.
\label{eq:app_attention_normalization}
\end{equation}
If every score is zero, the displayed attribution map is zero. The scores are mapped to the spatial grid of the \emph{actual decoder-visible image feature tokens}, accounting for visual token merging or spatial reordering. They are then bilinearly upsampled to the initial-image resolution and alpha-blended with $I_0$. Per-step normalization displays relative attribution within each panel.

\paragraph{Interpretation and scope.}
All panels within a trajectory use the same initial image, instruction, and static-prefix KV cache. Across steps, the GRU latent state $z_t$, its appended token $e_t=z_t$, the decoder readout, and the selected action can change. The diagnostic backward pass holds model parameters and the stored rollout state fixed.

Conditioned on fixed $z_t$ and prefix keys and values, this analysis describes the current decision's state-to-image-token readout through local sensitivity of the selected action logit.

Accordingly, $\widetilde r_t$ is a \emph{latent-state- and action-conditioned attribution map}. The first panel of each row shows the unchanged initial observation; subsequent panels display the attribution overlay and selected action.

\subsection{Qualitative Results across Tasks}

We next examine complete trajectories from FrozenLake, Maze, and MiniBehavior. These tasks respectively test whether visual attribution follows changing local hazards, wall-constrained topology, and phase-dependent object interactions while the visual input remains fixed.

\begin{figure}[!htbp]
\centering
\includegraphics[width=.75\textwidth]{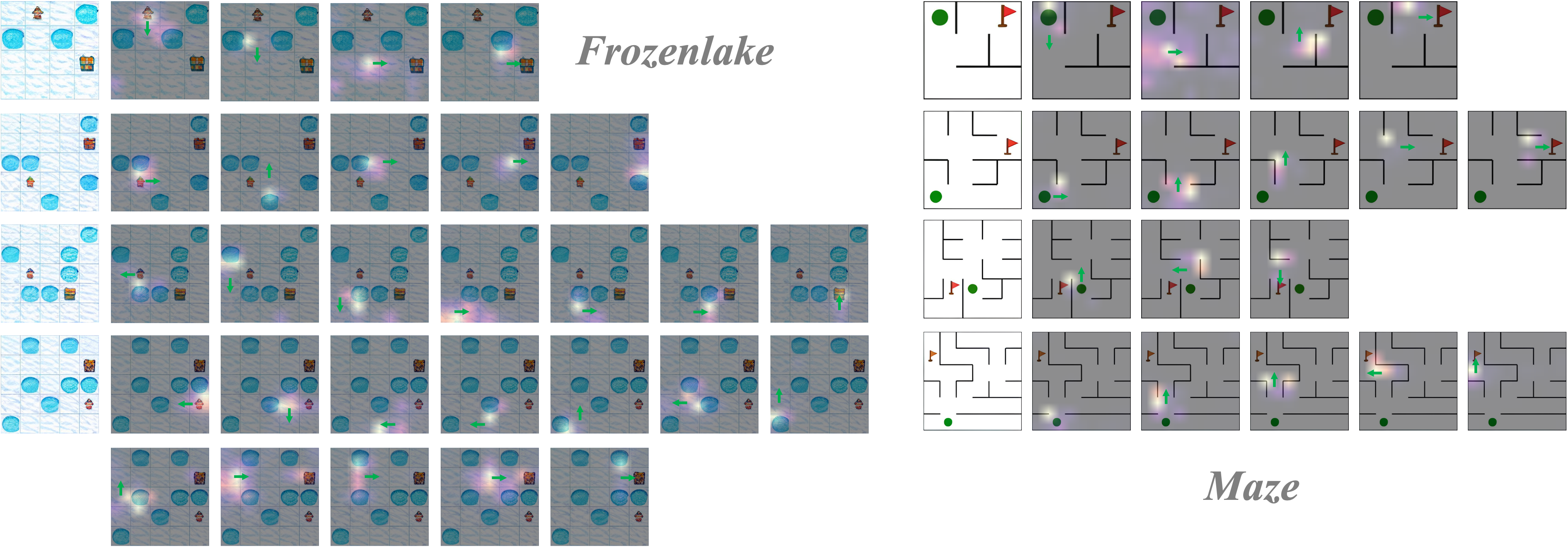}
\caption{Additional FrozenLake and Maze trajectories. In FrozenLake (left), attribution emphasizes nearby holes and safe local passages as the GRU latent state evolves with the predicted actions. In Maze (right), it moves across the fixed map and concentrates on relevant walls and openings. The shift across steps is consistent with latent-state-conditioned reading of a persistent scene.}
\label{fig:app_fro_maze_attention}
\end{figure}

Figure~\ref{fig:app_fro_maze_attention} contains examples with different map sizes, start locations, goals, and path lengths. In FrozenLake, high attribution frequently appears around hazards adjacent to the position reached by the predicted actions, which are relevant to avoiding an immediately failing transition. In Maze, attribution tracks the relevant wall segment or opening as the predicted plan progresses. Highlighted regions can include neighboring obstacles and corridors relevant to the current decision.

\begin{figure}[!htbp]
\centering
\includegraphics[width=0.65\textwidth]{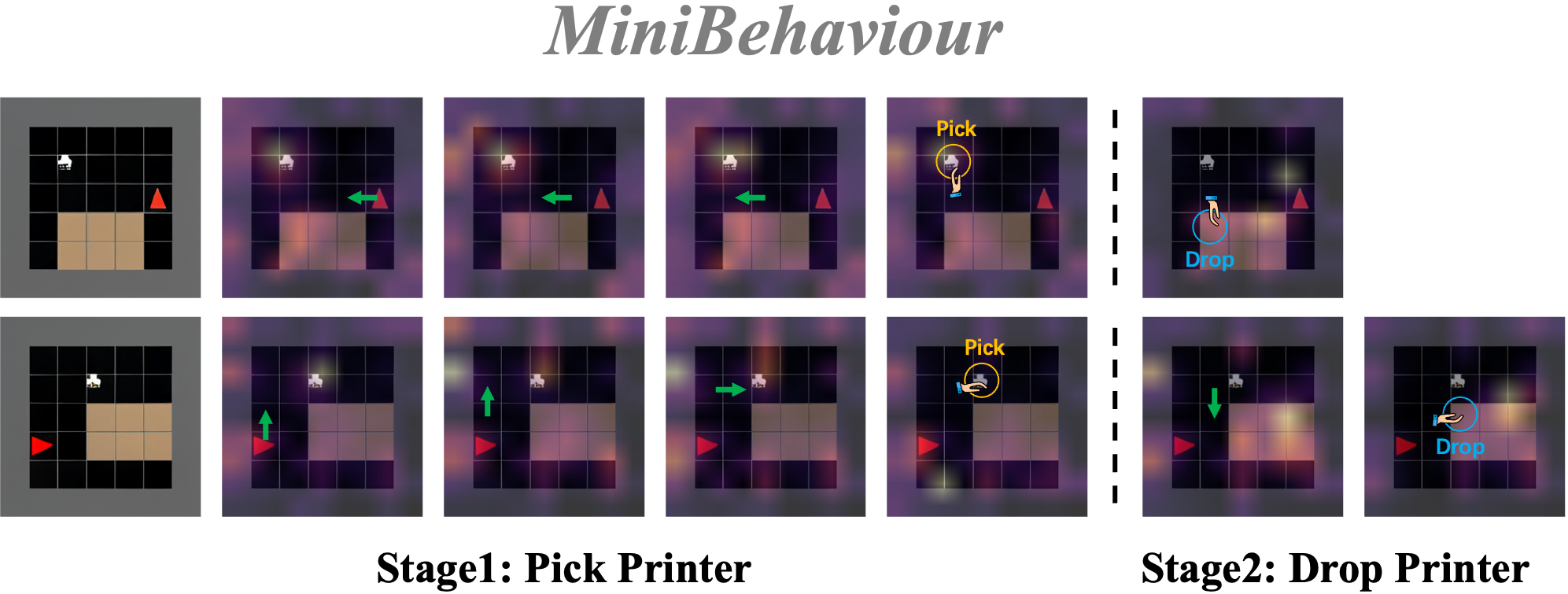}
\caption{Additional MiniBehavior trajectories. Each row follows a complete two-phase plan: the agent first navigates to the printer and executes \texttt{PICK}, then navigates to the table and executes \texttt{DROP}. The GRU updates its latent state from the predicted action history. Attribution correspondingly shifts from the printer and its approach path to the table and its approach path.}
\label{fig:app_mini_attention}
\end{figure}

MiniBehavior further tests whether attribution changes with the task phase represented implicitly through the action history. As shown in Figure~\ref{fig:app_mini_attention}, the pre-pick panels emphasize the printer and navigational constraints on the way to it. After \texttt{PICK}, the visual input remains fixed, and the GRU latent state is updated using the selected action; attribution then moves toward the table and relevant boundaries before \texttt{DROP}. This phase-dependent change provides qualitative evidence that the same scene is re-read according to the evolving latent state.

\section{Implementation and Experimental Details}
\label{sec:app_implementation}

\subsection{Task Interfaces and Data}

FrozenLake\footnote{\url{https://huggingface.co/yixu1/VPRL-7B-FrozenLake}} and Maze\footnote{\url{https://huggingface.co/yixu1/VPRL-7B-Maze}} use four actions, \texttt{UP}, \texttt{DOWN}, \texttt{LEFT}, and \texttt{RIGHT}. Their model state is a GRU latent vector updated from the action history. MiniBehavior\footnote{\url{https://huggingface.co/yixu1/VPRL-7B-MiniBehaviour}} additionally uses \texttt{PICK} and \texttt{DROP}, with carrying information modeled implicitly in the same recurrent state.

Initial-map tokens are decoded once with the MUSE VQ-VAE into $256\times256$ images and cached on disk; Qwen's native image processor then constructs the multimodal input. Rollout uses the initial observation and the recurrent state updated from predicted actions.

\begin{table}[!htbp]
\centering
\caption{Dataset-specific state and action interfaces.}
\label{tab:app_task_interfaces}
\begin{tabular}{lcc}
\toprule
Task & Model state & Action tokens \\
\midrule
FrozenLake & $z_t\in\mathbb{R}^{3584}$ & U/D/L/R \\
Maze & $z_t\in\mathbb{R}^{3584}$ & U/D/L/R \\
MiniBehavior & $z_t\in\mathbb{R}^{3584}$ & U/D/L/R/PICK/DROP \\
\bottomrule
\end{tabular}
\end{table}

\begin{apprevision}
\subsection{Datasets and Splits}
\label{sec:app_data}

To support a fair comparison with VPRL~\citep{vp}, we use its released training and test splits. No task instance appears in both sets. Each distinct physical layout is counted as one map within a split; different start or goal configurations on the same layout define different task instances. Table~\ref{tab:app_data_sizes} reports the numbers of task instances and unique maps. Supervised decision states are deduplicated within each training task by position and, for MiniBehavior, carrying status.

\begin{table}[!htbp]
\color{black}
\centering
\caption{Dataset sizes. Maps are distinct physical layouts within each split. Training samples and unique supervised decision states are counted before EOS fine-tuning.}
\label{tab:app_data_sizes}
\small
\setlength{\tabcolsep}{4pt}
\begin{tabular}{lrrrr}
\toprule
Task & \shortstack{Tasks\\train / test} & \shortstack{Unique maps\\train / test} & \shortstack{Training\\samples} & \shortstack{Unique supervised\\decision states} \\
\midrule
FrozenLake & 4,000 / 1,000 & 2,719 / 797 & 161,711 & 62,282 \\
Maze & 4,000 / 1,000 & 3,175 / 883 & 139,726 & 71,863 \\
MiniBehavior & 566 / 403 & 566 / 142 & 90,678 & 29,347 \\
\bottomrule
\end{tabular}
\end{table}

\paragraph{Task-specific safe actions.}
FrozenLake moves must stay within the grid and avoid holes; Maze moves must stay within the grid and respect walls. MiniBehavior moves follow traversability rules, while \texttt{PICK} requires an adjacent printer before pickup and \texttt{DROP} requires carrying the printer at an adjacent table position. Distance is the minimum remaining number of feasible actions to complete the task, including \texttt{PICK} and \texttt{DROP} for MiniBehavior. The safe set includes legal moves with known, unequal current and next-state distances, as well as goal-reaching moves, valid \texttt{PICK}, and successful \texttt{DROP}. The optimal subset includes moves that reduce distance by one and the same goal-reaching and interaction actions.

\end{apprevision}

\subsection{Optimization}

All main GRU-based models initialize the backbone from Qwen2.5-VL-7B-Instruct. The pretrained backbone weights are frozen and adapted using LoRA, while the GRU state module is learned jointly. LoRA is applied to the attention projections and MLP projections listed in Table~\ref{tab:app_training}. We use the soft target from the main paper with $\alpha=0.7$.

\begin{table}[!htbp]
\centering
\caption{Main training and evaluation configuration. The learning-rate scheduler is constructed for ten epochs, while training is stopped after epoch five.}
\label{tab:app_training}
\small
\renewcommand{\arraystretch}{0.85}
\begin{tabular}{ll}
\toprule
Setting & Value \\
\midrule
Backbone & Qwen2.5-VL-7B-Instruct \\
Trainable adaptation & LoRA + GRU state module \\
LoRA rank / \texttt{lora\_alpha} / dropout & 32 / 64 / 0.1 \\
LoRA target modules & \texttt{q/k/v/o\_proj}, \texttt{gate/down/up\_proj} \\
Optimizer & fused AdamW \\
Peak learning rate & $1.5\times10^{-4}$ \\
Schedule / warmup & cosine / 10\% of scheduled steps \\
Scheduled / executed epochs & 10 / 5 \\
Per-device batch size & 16 \\
Gradient accumulation & 1 \\
Number of training processes & 8 \\
Training hardware & 8$\times$ NVIDIA H800 GPUs \\
Effective global batch size & 128 \\
Training precision & bfloat16 mixed precision with TF32 enabled \\
Initialization / Trainer seed & 2026 / 42 \\
Checkpointing & once per epoch \\
Soft-target coefficient $\alpha$ & 0.7 \\
GRU input / hidden dimension & 3584 / 3584 \\
Image reconstruction resolution & $256\times256$ \\
Evaluation decoding & greedy over task action tokens \\
Evaluation environment dynamics & deterministic, no slip \\
\bottomrule
\end{tabular}
\end{table}

All models are trained on eight NVIDIA H800 GPUs with one distributed training process per GPU. The per-device batch size is 16 with one gradient accumulation step, giving an effective global batch size of 128. Training uses bfloat16 mixed precision, fused AdamW, a peak learning rate of $1.5\times10^{-4}$, and a cosine schedule with a 10\% warmup. The schedule is parameterized for ten epochs, and the reported checkpoint is saved after epoch five.

\begin{apprevision}

\section{Additional Experiments}
\label{sec:app_additional_experiments}

\subsection{History-Modeling Modules under the Same Supervision}
\label{sec:app_modules}

To isolate the contribution of the history-modeling module from that of supervision, we compare GRU, LSTM, and action-history Transformer updates using the same training data, soft-target supervision ($\alpha=0.7$), and optimization settings (Table~\ref{tab:app_modules}).

\begin{table}[!htbp]
\color{black}
\centering
\caption{History-modeling alternatives with the same training data and soft-target supervision. Entries report EM / PR (\%).}
\label{tab:app_modules}
\small
\begin{tabular}{lccc}
\toprule
Task & GRU & LSTM & Action-history Transformer \\
\midrule
FrozenLake & 99.50 / 99.55 & 98.00 / 98.81 & 97.50 / 98.95 \\
Maze & 96.30 / 97.98 & 96.30 / 97.49 & 92.50 / 95.25 \\
MiniBehavior & 83.87 / 90.56 & 73.70 / 85.96 & 66.25 / 84.49 \\
\bottomrule
\end{tabular}
\end{table}

The GRU achieves the strongest overall results, with the largest differences on MiniBehavior: 83.87\% EM versus 73.70\% for LSTM and 66.25\% for the action-history Transformer. On Maze, GRU and LSTM both reach 96.30\% EM, compared with 92.50\% for the Transformer.

All variants use 3,584-dimensional state tokens and action embeddings with the optimization settings in Table~\ref{tab:app_training}. The LSTM uses one cell, initializes its memory to zero, and passes its normalized hidden state to the decoder. The history Transformer uses one block with 28 attention heads, a $3584\rightarrow3584\rightarrow3584$ GELU feed-forward network, sinusoidal positions, and padding masks. At each update it processes all executed action embeddings followed by the current state, then normalizes the output at the final position. Both the GRU and Transformer update cores have 77,091,840 trainable parameters; the LSTM core has 102,789,120.

\subsection{State Capacity and Long-Horizon State Information Retention}
\label{sec:app_capacity}

\paragraph{State dimension.}
We test whether a compact recurrent state can support accurate planning by varying the GRU state dimension on Maze while holding the visual input and supervision fixed. A 1,792-dimensional state achieves 95.90\% EM and 97.98\% PR, matching the default state's PR with half the recurrent width (Table~\ref{tab:app_capacity}). Its complete state module has 32.14M parameters versus 89.96M for the default, demonstrating accurate planning with a compact dynamic representation.

\begin{table}[!htbp]
\color{black}
\centering
\caption{GRU state-dimension ablation on Maze with fixed visual input and supervision.}
\label{tab:app_capacity}
\begin{tabular}{lrr}
\toprule
GRU state dimension & EM (\%) & PR (\%) \\
\midrule
1,792 & 95.90 & 97.98 \\
3,584 (default) & 96.30 & 97.98 \\
7,168 & 90.80 & 95.70 \\
\bottomrule
\end{tabular}
\end{table}

For hidden dimension $d_h$, the initializer projects the pooled 3,584-dimensional scene embedding to $d_h$, and the learned state query, action embeddings, GRU, and state normalization use $d_h$ dimensions. A bias-free linear readout maps the state back to the decoder's 3,584 dimensions when $d_h\ne3584$; the default readout is the identity.

\paragraph{Long-horizon state information retention analysis.}
We examine whether the recurrent state preserves position and interaction information through successive updates along longer reference trajectories. We fit a linear probe to $z_t-z_0$ with SSVR frozen, using the original training tasks and a separate ridge regressor for each environment (intercept included, regularization coefficient 1). Evaluation covers 2,240 intermediate states from 318 held-out tasks with optimal reference plans of 7--14 actions, excluding initial and terminal states. Table~\ref{tab:app_state_probe} reports displacement recovery.

\begin{table}[!htbp]
\color{black}
\centering
\caption{Long-horizon state information retention analysis along optimal reference trajectories. Exact readout compares both displacement coordinates after nearest-integer rounding; MAE uses unrounded predictions in grid cells.}
\label{tab:app_state_probe}
\small
\begin{tabular}{lrr}
\toprule
Task & Exact readout (\%) & MAE (cells) \\
\midrule
FrozenLake & 99.363 & 0.02409 \\
Maze & 99.882 & 0.02426 \\
MiniBehavior & 99.676 & 0.02452 \\
\bottomrule
\end{tabular}
\end{table}

Displacement recovery exceeds 99.3\% on every task, with MAE below 0.025 grid cells. At history length six, accuracy ranges from 99.231\% to 100.000\%; MiniBehavior carrying-state recovery is 100.000\%, using a threshold of 0.5 on a third regression output. Through recurrent updates along longer reference trajectories, SSVR's recurrent state retains precise position and interaction information, demonstrating that a compact state can effectively represent the dynamic task information needed for multi-step execution.

\subsection{Initial-State Pooling}
\label{sec:app_initialization}

To examine SSVR's compatibility with different state-initialization strategies, we replace mean pooling with learnable-query attention pooling while retaining the same scene--state architecture. In the discrete experiments, $U_0$ in Equation~\ref{eq:gru_initialization} consists of the initial VQ scene tokens mapped through the backbone's input embedding layer; the continuous variant uses projected visual-tower features. In both cases, Qwen's visual tower encodes the image into the persistent prefix accessed at every decision.

Attention pooling introduces one learned query $q_{\mathrm{pool}}\in\mathbb{R}^{3584}$ and replaces the mean in the same initializer with
\[
c_0^{\mathrm{attn}}=\sum_p w_pU_{0,p},\qquad
w_p=\operatorname{softmax}_p\!\left(\frac{q_{\mathrm{pool}}^\top U_{0,p}}{\sqrt{3584}}\right).
\]
This adds 3,584 parameters. Table~\ref{tab:app_initialization} demonstrates SSVR's compatibility with different initialization strategies, with mean pooling providing a simple, effective default. Replacing it with learnable-query attention pooling further improves EM and PR by 1.10 and 1.20 percentage points, respectively, showing that the same scene--state architecture can benefit from stronger initialization modules.

\begin{table}[!htbp]
\color{black}
\centering
\caption{Initial-state pooling ablation on Maze.}
\label{tab:app_initialization}
\begin{tabular}{lrr}
\toprule
Initialization & EM (\%) & PR (\%) \\
\midrule
Mean pooling & 96.30 & 97.98 \\
Learnable-query attention pooling & 97.40 & 99.18 \\
\bottomrule
\end{tabular}
\end{table}

\subsection{Autonomous Termination}
\label{sec:app_eos}

We test whether SSVR can generate an optimal plan and stop upon task completion without being given the instance-specific shortest-solution length. We fine-tune each trained model for five additional epochs with EOS supervision to jointly learn action prediction and termination, then evaluate with a common maximum budget of 64 decisions, including EOS. Generation stops when EOS is emitted; exhausting the budget without EOS is recorded as no stop and counted as failure. The strict \emph{optimal plan and stop} metric requires a shortest successful action sequence followed immediately by EOS.

\begin{table}[!htbp]
\color{black}
\centering
\caption{Autonomous termination with a common 64-decision budget, including EOS. All values are percentages.}
\label{tab:app_eos}
\small
\setlength{\tabcolsep}{5pt}
\begin{tabular}{lrrrrr}
\toprule
Task & \shortstack{Optimal plan\\and stop} & \shortstack{Correct\\stop} & \shortstack{Early\\stop} & \shortstack{Late\\stop} & \shortstack{No\\stop} \\
\midrule
FrozenLake & 92.2 & 92.5 & 1.3 & 6.0 & 0.2 \\
Maze & 96.8 & 97.0 & 1.9 & 0.2 & 0.9 \\
MiniBehavior & 87.593 & 90.323 & 3.226 & 6.203 & 0.248 \\
\bottomrule
\end{tabular}
\end{table}

SSVR generates optimal plans and stops autonomously on 92.2\%, 96.8\%, and 87.593\% of FrozenLake, Maze, and MiniBehavior tasks, respectively.

EOS reuses the backbone's end-of-message token. Each training task contributes one terminal sample with an EOS target; nonterminal samples retain the action soft targets with zero EOS probability. All samples receive equal weight. Correct stopping requires EOS immediately after the first successful completion; optimal-plan-and-stop additionally requires a shortest successful plan. Early stopping emits EOS without completing the task; late stopping emits EOS after generating additional actions beyond the first successful completion. No stop means EOS is absent within the decision budget. All rates use all evaluation tasks as the denominator.

\subsection{Generalization beyond the Supervised History Length}
\label{sec:app_history_generalization}

To evaluate generalization beyond the supervised history range, we retrain SSVR with action prefixes restricted to at most five steps in the supervised decision-state samples. Evaluation uses 166 Maze tasks whose shortest plans require six to eight steps; a model trained with the full history range serves as the reference. Seven- and eight-action plans require decisions after prefixes longer than five steps, extending state updates beyond the supervised history range.

\begin{table}[!htbp]
\color{black}
\centering
\caption{Maze planning on 166 tasks requiring six to eight actions after restricting supervised action-prefix length.}
\label{tab:app_history_generalization}
\begin{tabular}{lrr}
\toprule
Training histories & EM (\%) & PR (\%) \\
\midrule
Restricted: action prefix $\leq 5$ steps & 78.313 & 92.958 \\
Full history range & 86.747 & 92.871 \\
\bottomrule
\end{tabular}
\end{table}

With supervision restricted to five-step prefixes, SSVR achieves 92.958\% PR and generates complete optimal plans on 78.313\% of the evaluated tasks. The full-history reference reaches 92.871\% PR and 86.747\% EM on the same subset.

\subsection{MazeFlip: Predictable View Changes without Intermediate Observations}
\label{sec:app_mazeflip}

MazeFlip evaluates whether SSVR can plan through predictable changes in view orientation using the initial image and action history. Actions are interpreted in the current view, mapped to world coordinates, and executed before updating the view: horizontal moves toggle a vertical reflection, and vertical moves toggle a horizontal reflection. The entire view is reflected, including walls, goal, and agent; successive reflections compose while world geometry remains fixed. The planner receives the initial unflipped image and tracks subsequent changes through its action-driven state.

The dataset is constructed from Maze and supplies 139,726 training decision samples and 1,000 test tasks, with 250 tasks at each grid size from $3\times3$ to $6\times6$. Training starts from Qwen2.5-VL-7B-Instruct with fresh LoRA adapters and a newly initialized state module, using the main training configuration.

After five epochs, SSVR reaches \textbf{86.20\% EM and 94.82\% PR}, demonstrating complete planning under predictable view changes.

\begin{figure}[!htbp]
\color{black}
\centering
\includegraphics[width=0.65\textwidth]{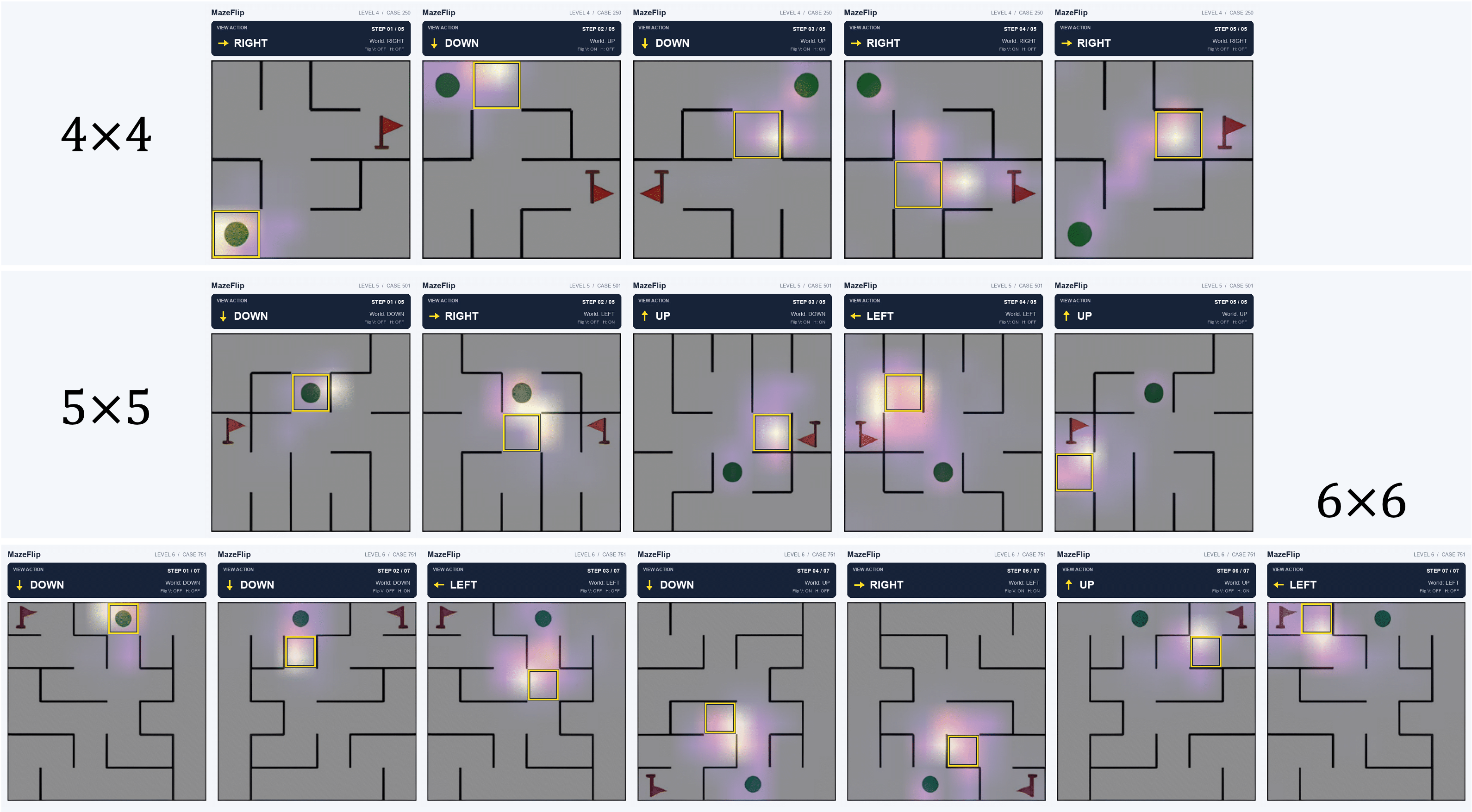}
\caption{Attention visualization for MazeFlip on $4\times4$, $5\times5$, and $6\times6$ mazes. Panels show successive decisions in the corresponding view coordinates; yellow boxes mark the current position, and headers indicate the predicted view action and its world-coordinate counterpart. The changing views illustrate the rollout; the model only receives the initial image.}
\label{fig:app_mazeflip_attention}
\end{figure}

Figure~\ref{fig:app_mazeflip_attention} shows attention shifting around the evolving position, nearby passages, and the goal approach as the view flips. For example, in the $4\times4$ trajectory, the second view action \texttt{DOWN} corresponds to world-coordinate \texttt{UP}. These examples illustrate state-conditioned visual reading that remains aligned with task progress across changing action coordinates.

\subsection{Continuous Navigation and General Action Interfaces}
\label{sec:app_continuous}

\paragraph{Continuous displacement prediction.}
This experiment tests whether the scene--state architecture extends from discrete action selection to continuous motion planning by adapting the action output and encoding interfaces. We use a unit-square scene with rectangular obstacles and a goal disk of radius 0.03. Inputs are $112\times112$ images. The data contain 5,000 training and 500 validation tasks, with independent layouts and a 20\% direct-path / 80\% detour composition. Expert paths are computed using a visibility graph and divided into displacements of length at most 0.08.

A $3584\rightarrow512\rightarrow2$ SiLU regression head predicts normalized displacement $u_t$, yielding $\Delta p_t=0.08u_t/\max(1,\|u_t\|_2)$. A $2\rightarrow512\rightarrow3584$ SiLU encoder maps $\Delta p_t/0.08$ to the GRU input. Training minimizes Smooth L1 loss ($\beta=1$) against normalized expert actions under reference histories, sampling eight decisions per task per epoch. We use AdamW, global batch size 32, and ten epochs: five at peak learning rate $1.5\times10^{-4}$ followed by five at $10^{-6}$, with validation-based checkpoint selection.

During evaluation, the planner generates from the initial image with a common 32-action budget. Success requires reaching the goal disk before a collision or boundary exit; collision checks cover each complete motion segment. For actual executed path length $L$ and shortest feasible length $L^\ast$ to the goal disk, path efficiency is
\[
\mathrm{PE} =
\begin{cases}
\dfrac{L^\ast}{\max\{L^\ast,L,10^{-9}\}}, & \text{if the plan succeeds},\\[4pt]
0, & \text{otherwise},
\end{cases}
\]
averaged over all evaluated instances and reported as a percentage. The reference $L^\ast$ is computed on the original obstacle geometry using a visibility graph and Dijkstra's algorithm. We report the epoch-7 checkpoint on the 128-task validation subset used for checkpoint selection, with 102 successful plans.

\begin{table}[!htb]
\color{black}
\centering
\caption{Two-dimensional continuous navigation on 128 validation tasks, using the initial image and a common 32-step budget.}
\label{tab:app_continuous}
\begin{tabular}{lrr}
\toprule
Evaluation & Success rate (\%) & Path efficiency (\%) \\
\midrule
Validation (128 tasks) & 79.69 & 77.04 \\
\bottomrule
\end{tabular}
\end{table}

Table~\ref{tab:app_continuous} demonstrates that the scene--state architecture can support continuous motion planning by adapting its action output and encoding interfaces.

\begin{figure}[!htb]
\color{black}
\centering
\includegraphics[width=0.65\textwidth]{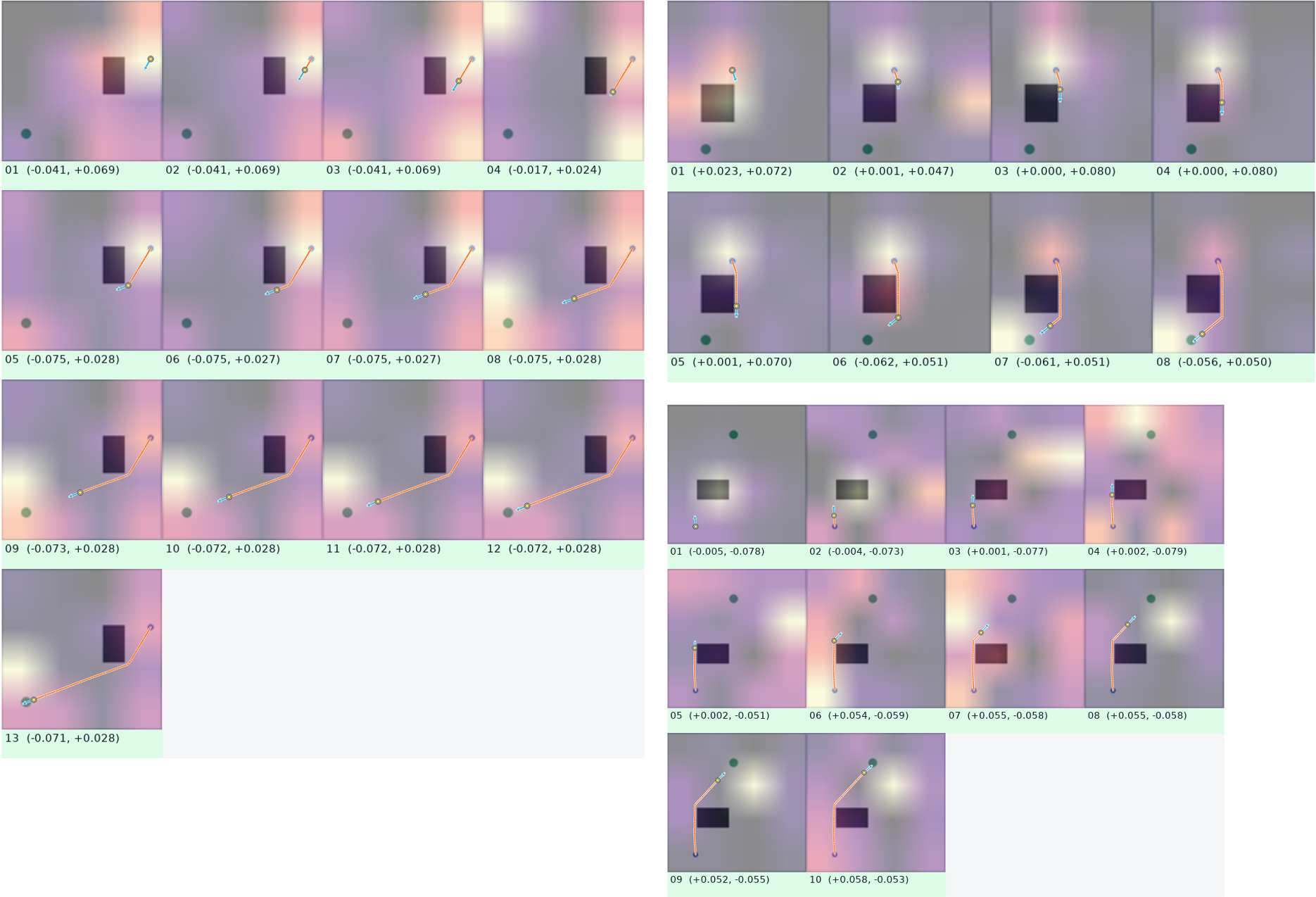}
\caption{Attention visualization for three continuous-navigation trajectories. Orange traces show the accumulated paths, arrows indicate the next predicted displacements, and green disks mark goals. Within each example, panels proceed from left to right and then to the next row; labels give the step and two-dimensional displacement.}
\label{fig:app_continuous_attention}
\end{figure}

Figure~\ref{fig:app_continuous_attention} shows obstacle-avoiding turns followed by goal-directed approaches. Attention shifts across obstacle-adjacent regions and the goal vicinity as planning progresses, illustrating how the recurrent state guides repeated reading of fixed scene geometry for continuous decisions.

\paragraph{Multi-token action interface extension.}
A possible extension for parameterized actions is to let the LM head generate an action type, parameter tokens, and an end-of-action marker. The completed action can then be encoded as a vector for one recurrent state update.

\section{Preprocessing, Prefix Construction, and Initialization Time}
\label{sec:app_full_timing}

We quantify the one-time cost of preparing reusable context and the initial state before action generation. Table~\ref{tab:app_full_timing} reports this setup cost for preprocessing, prefix-cache construction, and state initialization at different planned rollout lengths.

\begin{table}[!htbp]
\color{black}
\centering
\caption{One-time SSVR setup latency for different planned rollout lengths. Setup includes preprocessing, prefix-cache construction, and state initialization.}
\label{tab:app_full_timing}
\small
\begin{tabular}{lrrrr}
\toprule
Generated actions & 1 & 4 & 8 & 16 \\
\midrule
Setup time (ms) & 95.62 & 95.77 & 95.70 & 96.29 \\
\bottomrule
\end{tabular}
\end{table}

The setup cost remains approximately constant at 96~ms across the tested rollout lengths and is incurred only once per instance. Subsequent decisions reuse the cached context, with approximately constant per-step costs for action decoding and state updates. Total inference latency grows approximately linearly with rollout length, while the average setup cost per action decreases as the plan becomes longer.
\end{apprevision}

\begin{apprevision}
\section{Planning under Stochasticity and Partial Observability}
\label{sec:app_minibehavior_feedback}

Parkour offers an intuitive analogy: ``the brain plans, and the cerebellum executes.'' The environment does not pause for repeated analysis of the overall route; execution relies on practiced movements and rapid feedback adjustments. \emph{When online response time is insufficient for complex global decisions, the overall route must be established before acting}, so that rapid adjustments remain directed toward a clear goal.

In this division of labor, SSVR provides the pre-execution route and interaction goals, while a local navigator handles immediate deviations. We test whether SSVR can guide successful execution under stochastic motion and newly observed obstacles.

\paragraph{Task and perturbations.}
We use MiniBehavior tasks at difficulty level~6 and report the first 50 task--perturbation combinations in a seed-42 frozen evaluation schedule, covering 40 distinct held-out layouts. Each task requires picking up a printer and placing it on a table in a $6\times6$ scene, observed initially as a $256\times256$ image. Every episode combines two perturbations: each move attempts two cells with 20\% probability, stopping at obstructions, and a new obstacle appears along a generator-computed reference route. Obstacle placements preserve task reachability and allow local detours of at most eight cells. The initial image provides the static scene, while the newly introduced obstacle is revealed through local feedback only when it enters the robot's four-neighborhood.

\paragraph{Planning and local execution.}
All planning models receive the same VQ-reconstructed initial image and the same user task instruction, with method-specific system instructions specifying the output format and planning procedure. All methods share the route parser, local navigator $C$, and task--perturbation combinations. Predicted actions define waypoints and \texttt{PICK}/\texttt{DROP} checkpoints. The navigator requires no training and directly receives the robot's actual position and four-neighborhood traversability to follow the route, bypass obstacles, and correct slip-induced deviations. We reuse the existing Qwen2.5-VL-7B-based SSVR-GRU model without additional training. SSVR greedily generates its route before execution, stopping at \texttt{DROP} or 64 actions. Qwen4-Max either generates the full sequence in one request or predicts actions step by step from the initial image and its own predicted history, also before execution. Both Qwen variants use temperature~0 with thinking disabled. These three settings retain their initial plans during execution.

\paragraph{Evaluation and results.}
Success requires completing pickup and placement within 64 environment action calls, including failed attempts; a two-cell slip counts as one call. Table~\ref{tab:app_minibehavior_feedback} shows that SSVR with a fixed plan achieves 74\% success, compared with 4\% and 6\% for the two Qwen variants. These results demonstrate that SSVR-generated routes and interaction checkpoints support successful local execution under stochastic motion and locally revealed obstacles, achieving higher success than the two evaluated Qwen pre-execution planning strategies.

\paragraph{Zero-shot feedback adaptation.}
Building on this result, we further test zero-shot feedback adaptation. SSVR updates its remaining plan when no usable route is available, local detouring reaches eight movement calls, a loop is detected, or \texttt{PICK}/\texttt{DROP} fails. The initial image is retained, and the text prefix incorporates actual position, carrying status, local observations, action outcomes, execution history, and task coordinates and interaction regions. The recurrent state is reconstructed by replaying successful unit actions from the initial state; a two-cell slip replays the movement twice. The same action head then generates the remaining route for $C$, without parameter updates. With this training-free adaptation, success increases from 37/50 to 40/50, showing improved task completion when execution feedback is incorporated into remaining-plan generation.

\begin{table}[!htbp]
\color{black}
\centering
\caption{Main planning comparison and subsequent zero-shot feedback adaptation under slips and newly observed obstacles. All settings share navigator $C$.}
\label{tab:app_minibehavior_feedback}
\small
\begin{tabular}{lrr}
\toprule
Planning method (with navigator $C$) & Successes & Success rate (\%) \\
\midrule
SSVR: fixed plan & \textbf{37/50} & \textbf{74} \\
Qwen4-Max: full sequence & 2/50 & 4 \\
Qwen4-Max: step by step & 3/50 & 6 \\
\midrule
SSVR: zero-shot feedback adaptation & \textbf{40/50} & \textbf{80} \\
\bottomrule
\end{tabular}
\end{table}

Figure~\ref{fig:app_minibehavior_cases} illustrates five successful executions, including detours before and after pickup and recovery from repeated slips. Local corrections preserve the planned interaction goals and lead to successful placement.

\clearpage
\begin{figure}[p]
\color{black}
\centering
\resizebox*{!}{0.86\textheight}{%
\begin{minipage}{\linewidth}
\centering
\includegraphics[width=\linewidth,trim=47bp 155.5bp 47.5bp 155bp,clip]{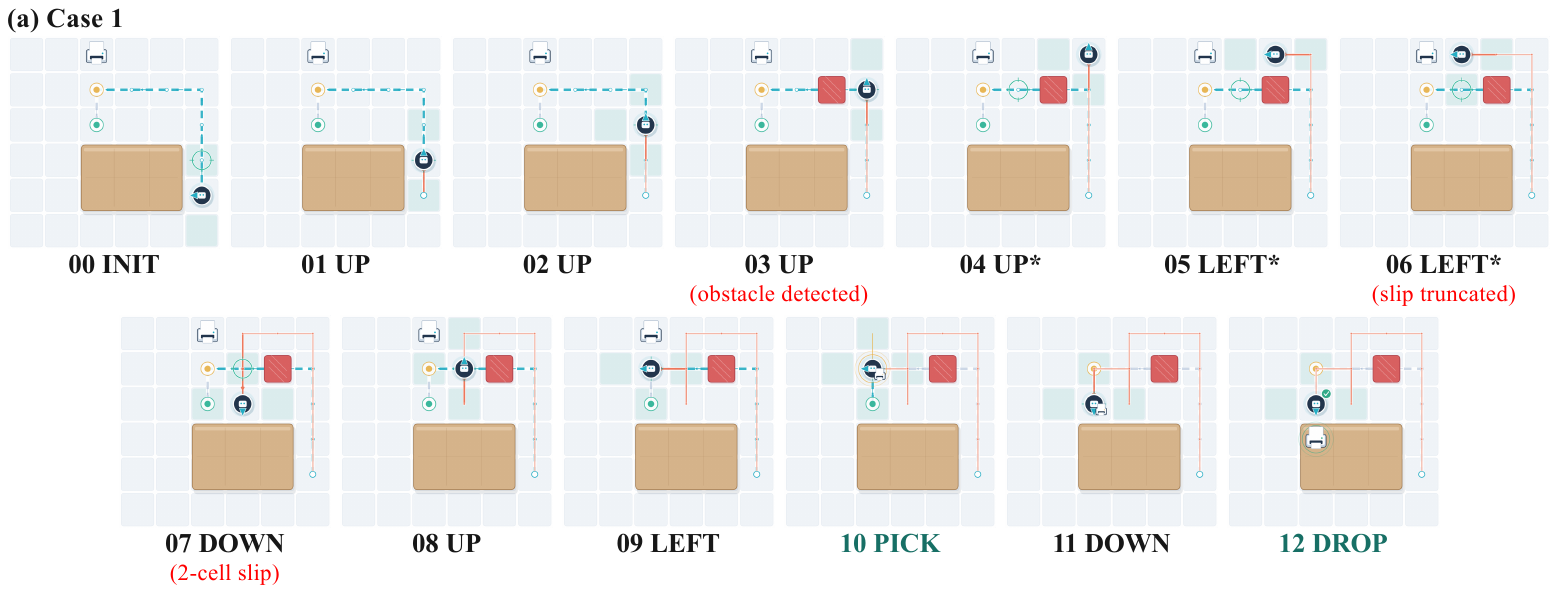}
\par\vspace{5pt}
\includegraphics[width=\linewidth,trim=47bp 147bp 48bp 143.5bp,clip]{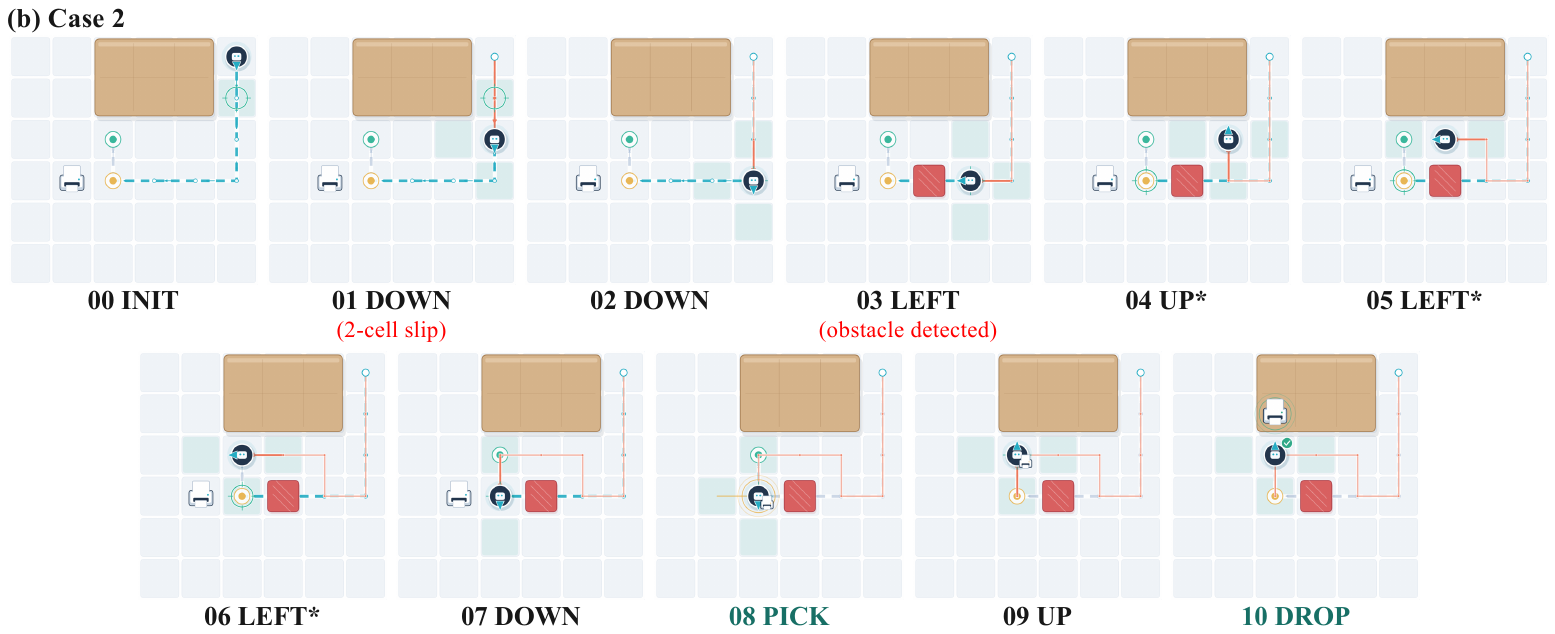}
\par\vspace{5pt}
\includegraphics[width=\linewidth,trim=47bp 121.5bp 48bp 118.5bp,clip]{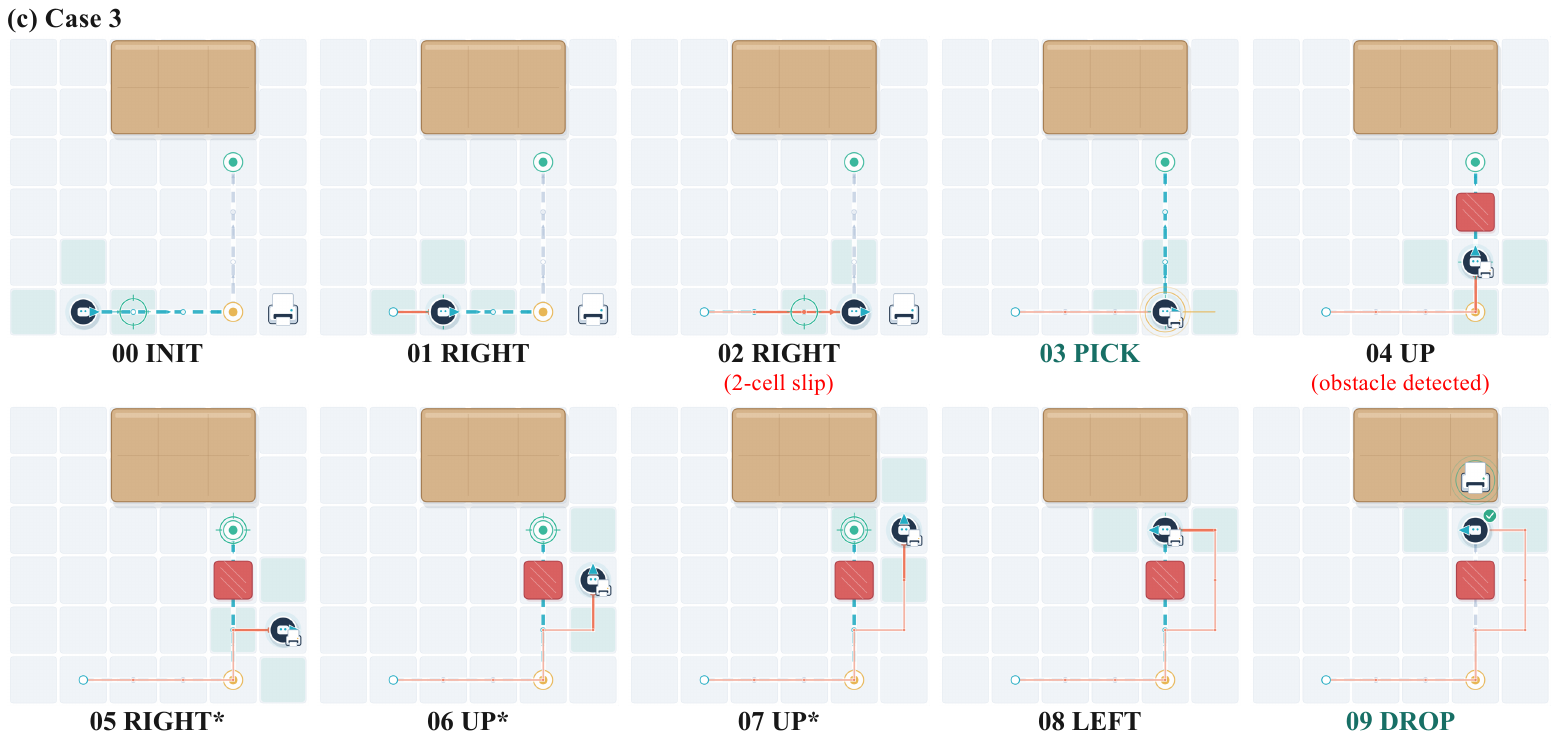}
\par\vspace{5pt}
\includegraphics[width=\linewidth,trim=47bp 131.5bp 48bp 131bp,clip]{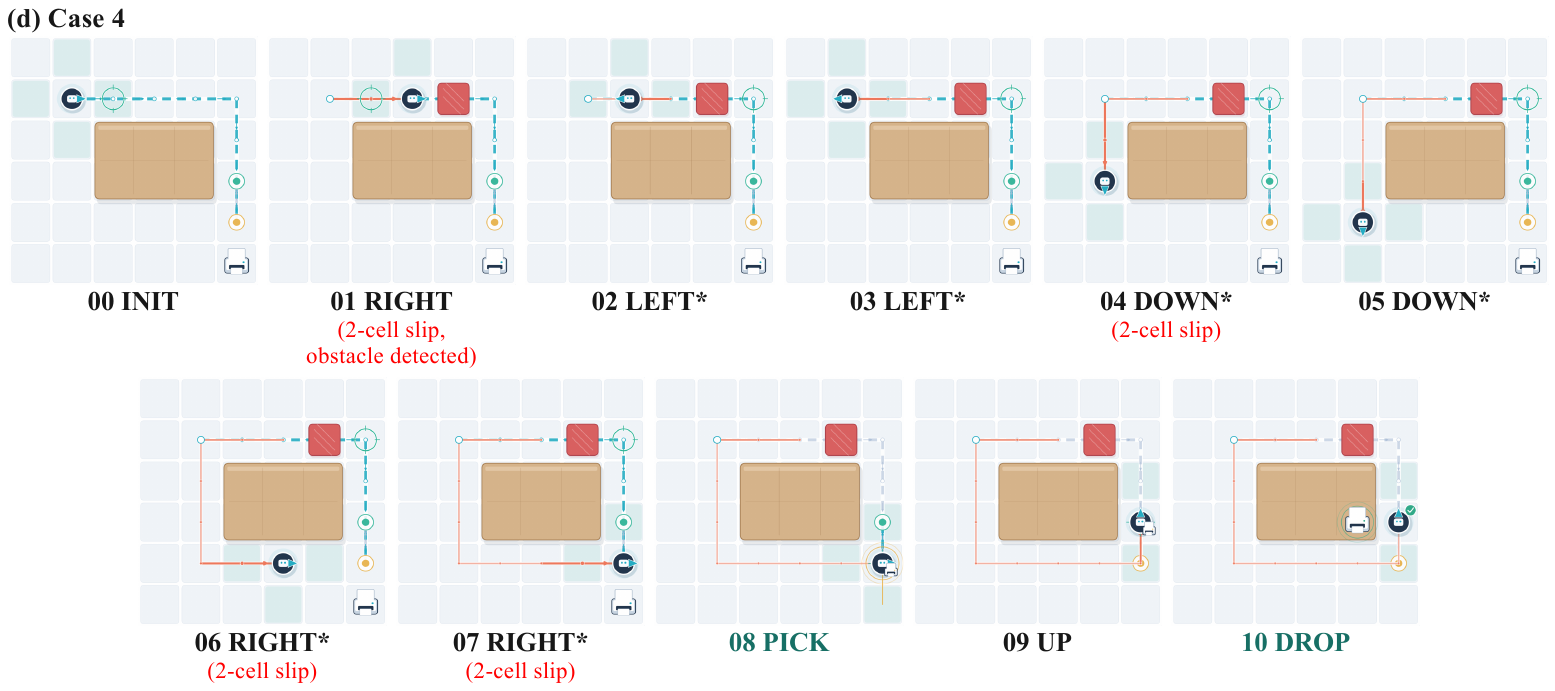}
\par\vspace{5pt}
\includegraphics[width=\linewidth,trim=47bp 147bp 48bp 143.5bp,clip]{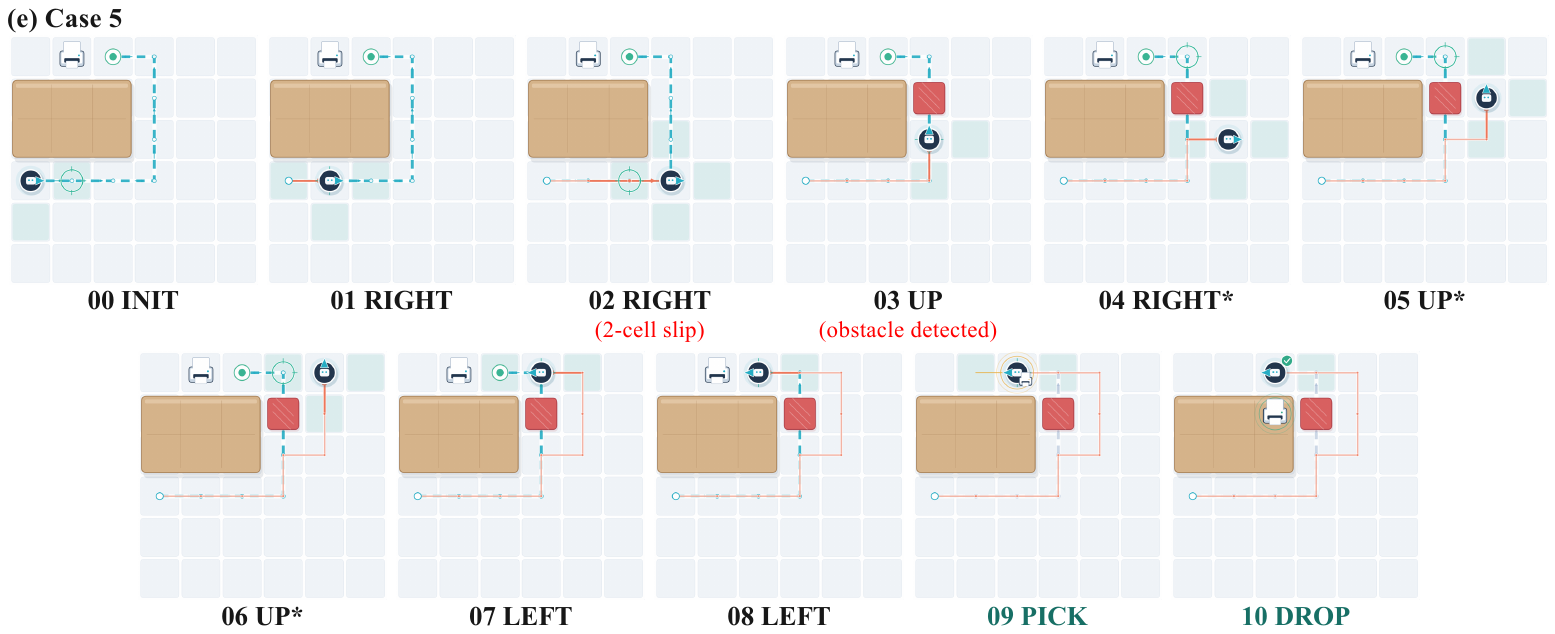}
\end{minipage}%
}
\caption{Complete execution sequences for five MiniBehavior cases. Cyan dashed lines denote planned routes, orange traces show executed motion, and red cells mark new obstacles. ${*}$ denotes a local detour. Slips and obstacle discovery are annotated.}
\label{fig:app_minibehavior_cases}
\end{figure}
\end{apprevision}